\documentclass[lettersize,journal]{IEEEtran}
\usepackage{amssymb,amsmath,amsfonts}
\usepackage{algorithmic}
\usepackage{algorithm}
\usepackage{array}
\usepackage{booktabs}
\usepackage{colortbl}
\usepackage[hidelinks]{hyperref}
\usepackage{graphicx}
\usepackage{makecell}
\usepackage[mathlines, switch]{lineno}
\usepackage{multirow}
\usepackage[sort&compress, numbers]{natbib}
\usepackage{pifont}
\usepackage{stfloats}
\usepackage[caption=false,font=footnotesize]{subfig}
\usepackage{textcomp}
\usepackage{url}
\usepackage{verbatim}
\usepackage[dvipsnames]{xcolor}

\hypersetup{
  colorlinks = true,
  urlcolor = CadetBlue,
  linkcolor = Cerulean,
  citecolor = Maroon
}

\definecolor{whitesmoke}{rgb}{0.96, 0.96, 0.96}

\begin{document}

\title{Choose Your Simulator Wisely: A Review on Open-source Simulators for Autonomous Driving}
\author{Yueyuan~Li, Wei~Yuan, Songan~Zhang, Weihao~Yan, Qiyuan~Shen, Chunxiang~Wang, and~Ming~Yang 
\thanks{This work is supported in part by the National Natural Science Foundation of China under Grants 62173228, 62203301, and 62203294. \textit{(Corresponding author: Ming Yang)}}
\thanks{Yueyuan Li, Weihao Yan, Qiyuan Shen, Chunxiang Wang, and Ming Yang are with the Department of Automation, Shanghai Jiao Tong University, Key Laboratory of System Control and Information Processing, Ministry of Education of China, Shanghai, 200240, CN (email: MingYANG@sjtu.edu.cn).}
\thanks{Wei Yuan is with the Innovation Center of Intelligent Connected Vehicles, Global Institute of Future Technology, Shanghai Jiao Tong University, Shanghai, 200240, CN}
\thanks{Songan Zhang is with the Global Institute of Future Technology, Shanghai Jiao Tong University, Shanghai, 200240, CN}
}

\markboth{}%
{}

\IEEEpubid{}

\maketitle

\begin{abstract}
    Simulators play a crucial role in autonomous driving, offering significant time, cost, and labor savings. Over the past few years, the number of simulators for autonomous driving has grown substantially. However, there is a growing concern about the validity of algorithms developed and evaluated in simulators, indicating a need for a thorough analysis of the development status of the simulators.
    
    To bridge the gap in research, this paper analyzes the evolution of simulators and explains how the functionalities and utilities have developed. Then, the existing simulators are categorized based on their task applicability, providing researchers with a taxonomy to swiftly assess a simulator's suitability for specific tasks. Recommendations for select simulators are presented, considering factors such as accessibility, maintenance status, and quality. Recognizing potential hazards in simulators that could impact the confidence of simulation experiments, the paper dedicates substantial effort to identifying and justifying critical issues in actively maintained open-source simulators. Moreover, the paper reviews potential solutions to address these issues, serving as a guide for enhancing the credibility of simulators.
\end{abstract}

\begin{IEEEkeywords}
Autonomous Driving, Simulator, Testing.
\end{IEEEkeywords}

\section{Introduction}

In the realm of autonomous driving, the importance of simulators has been widely acknowledged within academic and industrial communities. From 2022 to 2023, over 50\% of the methods published in this domain were either trained or tested in simulation environments. This trend is particularly pronounced in driving decision-making tasks, where over 70\% of publications underwent verification in simulators. Simulators, as compared to real-world scenarios, exhibit the unique capability to swiftly synthesize sensor data on a large scale, replicating extreme weather conditions, unpredictable behaviors of radical drivers, and intricate traffic scenarios. Moreover, simulators offer the unique advantage of replaying traffic accidents without posing safety hazards. Kalra et al. highlighted the impracticality of achieving statistical significance in autonomous vehicle performance validation through real-world testing alone, estimating a staggering 215 billion miles of testing would be required \cite{kalra2016driving}. Addressing this challenge, Shuo et al. presented compelling evidence that introducing proper simulation methodologies can accelerate the evaluation process $10^3$ to $10^5$ times faster \cite{feng2023dense}. Leading technology companies in the autonomous driving industry, including Waymo, Pony.ai, and Didi, emphasize the development and application of simulators \cite{waymo2021simulation, ponyai, didi}. Waymo's remarkable performance, reflected in the Disengagement Report from California between 2020 and 2023 \cite{sinha2021crash}, serves as compelling evidence of the critical role that realistic simulators play in achieving outstanding results \cite{waymo2021simulation}.

Open-source simulators are particularly meaningful in fostering collaboration within the researcher community. They significantly reduce entry barriers for individual researchers and small institutions, especially for those where hardware-based development is financially challenging and creating a new simulation platform is time-intensive \cite{dosovitskiy2017carla}. By doing so, open-source simulators alleviate researchers from the burden of repeatedly setting up training workflows and evaluation benchmarks, allowing them to concentrate more on algorithm development. Furthermore, open-source simulators facilitate resource-sharing by establishing standardized data formats and encouraging the sharing of sensor data on a common platform. This collaborative approach will result in significant time and cost savings in data collection efforts \cite{caesar2021nuplan}. Influential institutions such as the Computer Vision Center and Waymo have recognized the importance of open-source simulators, and thus, they are devoted to developing Carla and Wayax \cite{dosovitskiy2017carla, gulino2023waymax}. These endeavors underscore the growing acknowledgment of the value of open-source simulators in advancing research within the field of autonomous driving.

While several surveys have covered various aspects of autonomous driving, there is a distinct gap in the literature when it comes to a comprehensive exploration of simulators. Despite a considerable increase in open-source simulators in recent years, existing surveys in this domain primarily focus on available alternatives. Works such as \cite{zhou2022survey, holen2021evaluation, kaur2021survey} investigate the functionalities of a few well-known simulators, listing the specific traffic scenarios they support. There remains a critical need for an in-depth and systematic analysis of the evolution, taxonomy, and technical challenges autonomous driving simulators face.

The primary objective of this paper is to provide a thorough review of the development of autonomous driving simulators. The subsequent sections are organized as follows:

Section \ref{related-works} reviews existing literature in the autonomous driving field, covering surveys that mention simulators, reviews specific to simulators for autonomous driving, and works discussing issues in simulators for autonomous driving.

Section \ref{history} overviews the evolution of simulators over the past three decades, proposing a split into the incipient, dormant, and outbreak periods based on development patterns. A tendency to develop open-source simulators and expand simulation task scope has been identified from the development history.

Section \ref{category} introduces a taxonomy categorizing simulators based on their suitability for specific tasks. The influential simulators are classified into five categories: traffic flow simulator, sensory data simulator, driving policy simulator, vehicle dynamics simulator, and comprehensive simulator.  Each category is explained, common functionalities are detailed, and guidelines for selecting appropriate tools are provided.

Section \ref{issues} identifies critical issues in open-source simulators, justifying the need to address them and warning about potential hazards that could compromise the validity of simulation-based experiments. This section summarizes potential methodologies to improve the simulators' performance about each identified issue.

\section{Related Works}
\label{related-works}

Most surveys and reviews related to autonomous driving focused on specific tasks \cite{gonzalez2015review, paden2016survey, feng2021review}, algorithm categories \cite{grigorescu2020survey, haydari2020deep, cui2021deep}, and overall system architecture \cite{yurtsever2020survey, tampuu2020survey}. These reviews admitted that simulators have a wide application in various scenarios. Yurtsever et al. demonstrated the capability of simulators to generate diverse data that captures different weather conditions \cite{yurtsever2020survey}. This way of data augmentation enhances the robustness of perception models, enabling them to perform effectively in a variety of scenarios \cite{johnson2016driving}. Similarly, Grigorescu et al. emphasized the importance of simulation for deep learning-based driving decision-making models, as these models require diverse interaction behavior to improve their decision-making capabilities \cite{grigorescu2020survey}. This viewpoint is echoed in other reviews focusing on decision-making and motion-planning algorithms \cite{katrakazas2015real, schwarting2018planning, claussmann2019review, gidado2020survey}. What's more, Tampuu et al. expressed concerns about the fidelity of simulators, highlighting the challenges of transferring decision models trained in virtual environments to the real world \cite{tampuu2020survey}. As simulators are not the primary focus of these surveys and reviews, most of them merely provide a brief list of available simulation platforms related to their respective themes \cite{wei2021autonomous}.

It was not until 2020, with the increasing number of open-source simulators, that reviews and discussions focusing on simulators began to emerge. Zhou et al. conducted a detailed survey of seven popular simulators, including CARLA, LGSVL, CarSim, AirSim, Prescan, Matlab/Simulink, and CarMaker. Their work explored the functionalities of these simulators and summarized their respective application scopes \cite{zhou2022survey}. Holen et al. conducted a similar study but expanded the range of simulators, offering a broader perspective on the available options \cite{holen2021evaluation}. The simulators mentioned in Kaur et al.'s research were mostly consistent with Zhou's work, with the main difference being the replacement of AirSim with Gazebo \cite{kaur2021survey}. This survey qualitatively describes the performance of the simulators, along with a summary of the desired functionalities and characteristics of simulators. The existing reviews intend to assist researchers in choosing a suitable simulator for their task, so their main content is to investigate the alternatives and check the available functionalities.

Recently, the discussion about the limitations of simulators is gaining attention. Hu et al. stressed the need to narrow the gap between simulators and the real world by automating the reconstruction of simulation scenarios and improving the sim2real model transfer efficiency \cite{hu2023sim2real}. \cite{ding2023survey} holds the same opinion and thoroughly reviews the potential solutions for generating driving scenarios. Meanwhile, Zhang et al. identified that enhancing the intelligence of traffic participants' behaviors is a promising avenue to boost simulator performance \cite{zhang2022trajgen}. Similarly, Siebke et al. emphasized the importance of including human errors in the traffic simulation process to enhance realism \cite{siebke2022traffic}. While these existing works successfully pinpoint critical issues within simulators, they tend to focus on isolated specific challenges, creating a void in comprehensively cataloging all the urgent problems that persist in the ongoing development of simulators.

This paper differs from previous works by introducing several distinctive features. It clarifies the relationship between the research tasks and the simulator categories in the ADS, offering task-specific recommendations for open-source simulators. Moreover, the paper conducts a comprehensive analysis of critical issues in existing simulators, accompanied by a compelling rationale emphasizing the urgency for their resolution. In addition to identifying these issues, the paper summarizes the potential solutions, providing valuable insights for addressing the identified challenges.

\section{History of Simulators}
\label{history}

\begin{figure*}
    \centering
    \includegraphics[width=\linewidth]{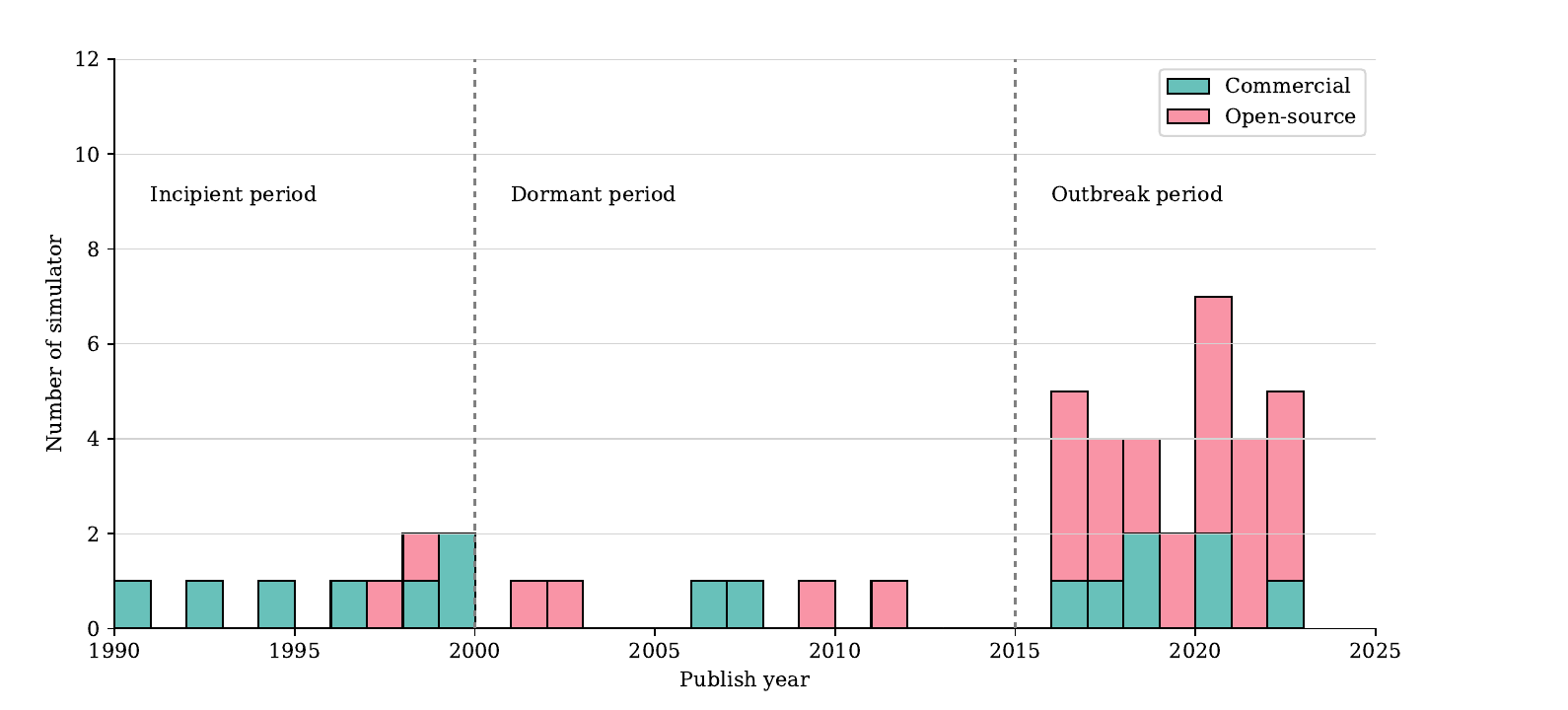}
    \caption{The number of simulators released by year.}
    \label{simulator-history}
\end{figure*}

The history of simulators for autonomous driving provides valuable insights into the technological evolution within this domain. By observing the number of simulators released every year (Fig. \ref{simulator-history}), this paper suggests categorizing the development of simulators into three distinct periods: the incipient period, the dormant period, and the outbreak period.

The era before 2000 is characterized as the incipient period, during which simulators began to be deployed to improve the intelligence of vehicles and the transportation system. Notable commercial entries during this time included PTV Vissim by PTV Group in 1992, concentrating on simulations related to traffic flow and logistics system \cite{PTVvissim}, and Paramics in 1994, offering comparable functionalities \cite{smith1995paramics}. Complementing these commercial options, the open-source simulator SUMO emerged as a pivotal force in traffic flow simulation \cite{lim2017sumo}. Simultaneously, in the realm of vehicle dynamics, CarSim and IPG CarMaker, established in the 1990s, forged enduring partnerships with various vehicle manufacturers \cite{carsim, carmaker}. Their reliance on extensive real-world data contributed to their dominance in this domain. RFpro, entering the scene in 2007, stood out as one of the few simulators matching the vehicle dynamics fidelity of CarSim and IPG CarMaker, thanks to real-world data collection from racing cars \cite{rfpro}. During the incipient period, the primary focus of simulators was on two topics: traffic flow and vehicle dynamics. These simulators were developed under well-developed theories, operated on explicit rules and physics models, and demanded relatively modest computational resources.

\begin{figure*}
    \centering
    \includegraphics[width=0.9\linewidth]{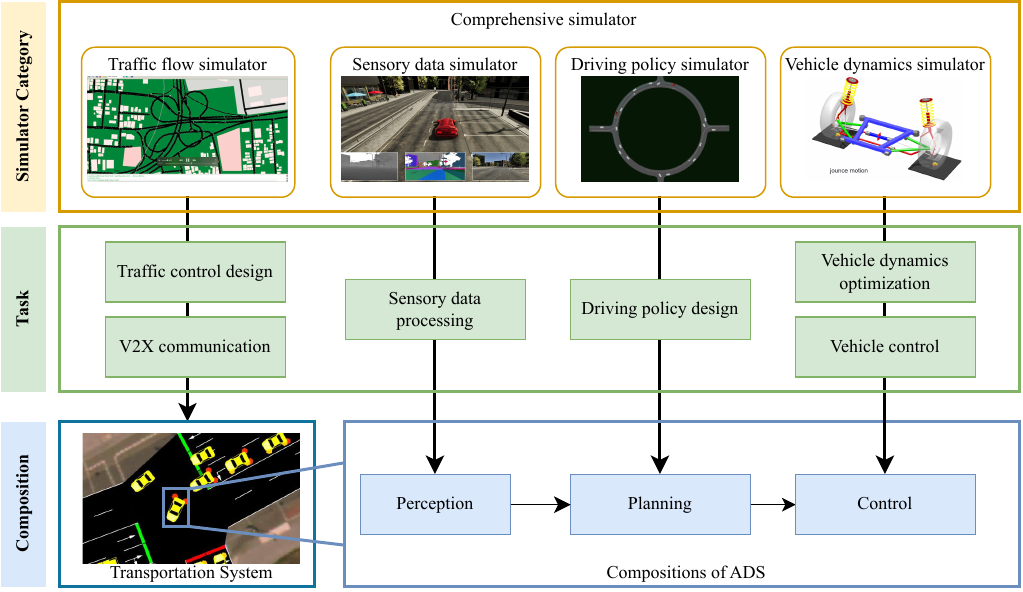}
    \caption{The relationship between the simulator categories, tasks in the ADS, and the compositions of the ADS \cite{SUMO2018, carsim, shah2018airsim, althoff2017commonroad}.}
    \label{simulator-category}
\end{figure*}

Between 2000 and 2015, the period witnessed a dormant phase in simulator development, primarily attributed to technological bottlenecks. Shortages in computational resources hindered support for high-performance and large-scale computations. Essential software techniques, including rendering, 3D modeling, and physics engines, were still under development. These factors collectively made the creation of high-fidelity simulators a costly endeavor. Furthermore, the potential applications of simulators remained underexplored during that time. Learning-based methods had not yet demonstrated their potential, resulting in a lack of urgent demand for generating extensive high-fidelity data. These technological barriers and a dearth of compelling motivations led to a decade-long gap. Only a select few simulators survived and left a lasting impact. Among them were PreScan, supported by Siemens with funding and hardware assistance \cite{prescan}; rFpro, which benefited from financial backing and data support from Formula 1 (F1) racing groups \cite{rfpro}; and VI-grade, a comprehensive simulation project facilitated by HBK Investment \cite{vigrade}. The survival of these simulators relied on substantial investments from world-leading profit organizations.

A significant growth in the number of simulators began in 2016. Thus, this paper named this the outbreak period. The advent of Neural Networks (NN) boosted the performance of models in many ADS tasks. Learning-based algorithms depend heavily on large-scale data. An image dataset sourced from Grand Theft Auto V (GTA V) underscored the simulators' advantage in synthesizing diverse sensory data \cite{martinez2017beyond}. Moreover, simulators provided an ideal environment for training and testing end-to-end ADS, where the costs for collisions and failures are negligible compared to real-world scenarios \cite{wang2019end}. Despite the termination of the interface for ADS development by GTA V's manufacturer, the research community embarked on developing independent simulators striving to approximate the real world as much as possible. Microsoft's Airsim, LG's SVL, and CVC's CARLA emerged, designed to generate diverse sensory data and simulate the movement of self-driving vehicles \cite{shah2018airsim, rong2020lgsvl, dosovitskiy2017carla}. With an awareness of the computation and time cost of high-fidelity simulators, a trend emerged favoring only providing 2D BEV semantic information to train and verify driving policies \cite{highway-env, althoff2017commonroad, caesar2021nuplan}.

The evolution history of simulators for autonomous driving reveals some trends:

\begin{itemize}
    \item \textbf{Rise of open-source simulators:} The increasing prevalence of open-source simulators might be explained by two factors. Firstly, technological advancements have made the development of simulators more accessible. The decreasing costs and enhanced availability of computing resources and software tools have made it easier for developers to create and distribute open-source simulators. Secondly, open-source simulators would benefit the entire autonomous driving industry. These platforms might contribute to establishing standardized data formats and evaluation benchmarks. This, in turn, fosters a more cohesive and efficient research environment, facilitating the sharing and validation of findings across the industry.

    \item \textbf{Diversification of task scope:} In the incipient and dormant periods, simulators primarily focused on examining traffic control designs and vehicle structures. However, as the focus in autonomous driving research shifted towards vehicle-centric exploration, simulators assumed a crucial role in the training and testing of algorithms. The expansion in application range has engendered a higher demand for simulation quality, especially in terms of fidelity and diversity.

\end{itemize}

\section{Categorization of Simulators}
\label{category}

\begin{table*}[htb]
    \caption{Basic information of categorized simulators. Functions: \ding{172} Traffic control design; \ding{173} V2X communication; \ding{174} Sensor data processing; \ding{175} Driving policy design; \ding{176} End-to-end driving policy design; \ding{177} Vehicle dynamics optimization; \ding{178} Vehicle control}
    \label{simulator-overview}
    \centering
    \begin{tabular}{l | l c c c l} \toprule[2pt]
        \textbf{Category} & \textbf{Simulator} & \textbf{Released Year} & \textbf{Active Maintenance} & \textbf{Open-source} & \textbf{Related Tasks} \\ \midrule[1pt]
        \multirow{7}{*}{Traffic flow simulator} & PTV Vissim \cite{PTVvissim} & 1992 & $\surd$ & - & \ding{172} \\
        & Paramics \cite{smith1995paramics} & 1994 & $\surd$ & - & \ding{172} \\
        & Aimsun \cite{barcelo2005dynamic} & 1999 & $\surd$ & - & \ding{172} \ding{173} \\
        & SUMO \cite{SUMO2018} & 2001 & $\surd$ & $\surd$ & \ding{172} \ding{173} \\
        & POLARIS \cite{auld2016polaris} & 2016 & $\surd$ & - & \ding{172} \ding{173} \\
        & Flow \cite{wu2017flow} & 2017 & - & $\surd$ & \ding{172} \\
        & CityFlow \cite{zhang2019cityflow} & 2019 & - & $\surd$ & \ding{172} \\ \midrule

        \multirow{5}{*}{Sensory data simulator} & Sim4CV \cite{muller2018sim4cv} & 2016 & - & - & \ding{174} \ding{178} \\
        & AirSim \cite{shah2018airsim} & 2017 & - & $\surd$ & \ding{174} \ding{175} \\
        & Parallel Domain \cite{paralleldomain} & 2017 & ? & - & \ding{174} \\
        & SVL \cite{rong2020lgsvl} & 2018 & - & $\surd$ & \ding{174} \ding{175} \\
        & UniSim \cite{yang2023unisim} & 2023 & $\surd$ & - & \ding{174} \\
        \midrule

        \multirow{18}{*}{Driving policy simulator}
        & TORCS \cite{wymann2000torcs} & 1997 & - & $\surd$ & \ding{176} \\
        & VDrift \cite{kehrle2011optimal} & 2011 & $\surd$ & $\surd$ & \ding{176} \\
        & CarRacing \cite{carracing} & 2016 & $\surd$ & $\surd$ & \ding{175} \\
        & Udacity \cite{udacity} & 2016 & - & $\surd$ & \ding{176} \\
        & CommonRoad \cite{althoff2017commonroad} & 2017 & $\surd$ & $\surd$ & \ding{175} \\
        & highway-env \cite{highway-env} & 2018 & $\surd$ & $\surd$ & \ding{175} \\
        & MACAD \cite{palanisamy2020multi} & 2019 & ? & $\surd$ & \ding{175} \\
        & BARK \cite{bernhard2020bark} & 2020 & ? & $\surd$ & \ding{175} \\
        & DriverGym \cite{kothari2021drivergym} & 2020 & - & $\surd$ & \ding{175} \\
        & SMARTS \cite{zhou2020smarts} & 2020 & $\surd$ & $\surd$ & \ding{175} \ding{176} \\
        & SUMMIT \cite{cai2020summit} & 2020 & ? & $\surd$ & \ding{175} \ding{176} \\
        & DI-Drive \cite{didrive} & 2021 & ? & $\surd$ & \ding{176} \\
        & L2R \cite{herman2021learn} & 2021 & $\surd$ & $\surd$ & \ding{176} \\
        & MetaDrive \cite{kar2019meta} & 2021 & $\surd$ & $\surd$ & \ding{175} \ding{176} \\
        & NuPlan \cite{caesar2021nuplan} & 2021 & $\surd$ & $\surd$ & \ding{175} \\
        & InterSim \cite{sun2022intersim} & 2022 & $\surd$ & $\surd$ & \ding{175} \\
        & Nocturne \cite{vinitsky2022nocturne} & 2022 & ? & $\surd$ & \ding{175} \\
        & TBSim \cite{xu2023bits} & 2023 & $\surd$ & $\surd$ & \ding{175} \\
        & Waymax \cite{gulino2023waymax} & 2023 & $\surd$ & $\surd$ & \ding{175} \\ \midrule

        \multirow{6}{*}{Vehicle dynamics simulator} & CarSim \cite{carsim} & 1996 & $\surd$ & - & \ding{177} \ding{178} \\
        & Webots \cite{michel2004cyberbotics} & 1998 & $\surd$ & $\surd$ & \ding{177} \ding{178} \\
        & CarMaker \cite{carmaker} & 1999 & $\surd$ & - & \ding{174} \ding{177} \ding{178} \\
        & Gazebo \cite{koenig2004design} & 2002 & ? & $\surd$ & \ding{177} \ding{178} \\ 
        & VI-CarRealTime \cite{vigrade} & 2009 & $\surd$ & - & \ding{177} \ding{178} \\ 
        & Matlab \cite{matlab} & 2018 & $\surd$ & - & \ding{177} \ding{178} \\ \midrule

        \multirow{9}{*}{Comprehensive simulator} & SCANeR Studio \cite{scaner} & 1990 & $\surd$ & - & \ding{174} \ding{175} \ding{176} \ding{177} \ding{178} \\
        & Virtual Test Drive \cite{vtd} & 1998 & $\surd$ & - & \ding{174} \ding{175} \ding{176} \ding{177} \ding{178} \\
        & PreScan \cite{prescan} & 2006 & $\surd$ & - & \ding{174} \ding{175} \ding{176} \ding{177} \ding{178} \\
        & rFpro \cite{rfpro} & 2007 & $\surd$ & - & \ding{174} \ding{176} \ding{177} \ding{178} \\
        & CARLA \cite{dosovitskiy2017carla} & 2016 & $\surd$ & $\surd$ & \ding{174} \ding{175} \ding{176} \ding{178} \\
        & DeepDrive \cite{Deepdrive} & 2018 & ? & $\surd$ & \ding{174} \ding{175} \ding{176} \ding{177} \\
        & Nvidia Drive Sim \cite{nvidia2020drivesim} & 2020 & $\surd$ & - & \ding{174} \ding{176} \ding{177} \ding{178} \\
        & Vista \cite{amini2022vista} & 2020 & $\surd$ & $\surd$ & \ding{174} \ding{176} \\
        & VI-WorldSim \cite{vigrade} & 2020 & $\surd$ & - & \ding{174} \ding{175} \ding{176} \ding{177} \ding{178} \\
        \bottomrule[2pt]
    \end{tabular}
\end{table*}

This section classifies simulators based on their compatibility with distinct ADS tasks. Most simulators can be categorized into five groups: traffic flow simulator, vehicle dynamics simulator, sensory data simulator, driving policy simulator, and comprehensive simulator. It is important to note that certain simulators may exhibit some overlap in functionalities. Each subsection in this section will delineate a specific category and provide practical recommendations for selecting actively maintained open-source simulators for associated tasks.

Table \ref{simulator-overview} presents a comprehensive overview of simulators, including their release year, maintenance status, open status, and relevance to ADS tasks. This paper identifies open-source simulators as those with source codes openly available online. Simulators that distribute their core for free but may charge for technical support or customized functionalities are also considered open-source \cite{michel2004cyberbotics, dosovitskiy2017carla}. Commercial simulators are covered when discussing the history and taxonomy of the simulators, given their dominant position in certain categories. In the \textit{Active Maintenance} column, those simulators without an update in the past year are annotated with a question mark. A simulator is considered no longer actively maintained if it has been out of maintenance for over one year. Considering the proliferation of simulators in recent years, this paper strives to cover relatively influential simulators, defined as those with over 100 citations on Google Scholar or over 100 stars on code-sharing platforms like Github. The organization of the table adheres to the proposed taxonomy. A detailed exploration of the functionalities of actively maintained open-source simulators will be presented in the subsequent subsections.

\subsection{Traffic Flow Simulator}

Traffic flow simulators are software that can simulate the movement of vehicles and other dynamic traffic participants within a transportation system. Typically, a traffic flow simulator may include the following features:

\begin{itemize}
    \item Road network customization: A traffic flow simulator enables users to customize the road network by providing detailed information on traffic rules, including speed limits, traffic lights, and road markings. Users can edit lanes, junctions, and traffic rules to tailor the simulation to specific scenarios.
    \item Microscopic traffic simulation: This aspect involves modeling individual vehicles and their interactions. Each vehicle is simulated independently, considering factors such as acceleration, lane following, and lane changing. To accommodate city-level traffic flow simulation, these simulators simplify the behavior models of vehicles.
    \item Visualization: Traffic flow simulators tend to provide a simplified bird-eye-view (BEV) graphical visualization, illustrating the location and status of traffic participants across the map.
\end{itemize}

Initially, traffic flow simulators were employed to gain insights into urban mobility, road infrastructure planning, and traffic control strategy design \cite{PTVvissim, smith1995paramics, wu2017flow, zhang2019cityflow}. Recent advancements include expanding their support to the development of Connected Autonomous Vehicles (CAV) and Vehicle-to-Everything (V2X) technologies \cite{barcelo2005dynamic, auld2016polaris, SUMO2018}, reflecting a growing attention to roadside cooperation perception for autonomous driving \cite{olaverri2018connection}.

SUMO stands out as the only actively maintained open-source option among the available traffic flow simulators. It supports a wide range of map formats and keeps improving its algorithms and hardware efficiency. These together make it the recommended choice when working on tasks like transportation planning, traffic control design, road infrastructure design, and V2X and CAV communication.

\subsection{Sensory Data Simulator}

Sensory data simulators are software that can generate highly realistic sensory data and ground truth for perception tasks. To support multi-modal perception tasks, a sensory data simulator must be capable of providing synthesis results from various sensors. One distinct advantage of sensory data simulators is their capacity to expedite data collection under challenging conditions, particularly adverse weather scenarios, thereby saving valuable time \cite{kalra2016driving}. Table \ref{simulator-sensor}-\ref{simulator-weather} outlines the sensors, ground truth annotations, and adverse weathers supported by each sensory data simulator. CARLA is included as it is capable of simulating high-fidelity sensory data.

\begin{table}[htb]
    \centering
    \caption{A comparison of the sensors that actively maintained open-source simulators can synthesize.}
    \label{simulator-sensor}
    \begin{tabular}{l | c c c c c} \toprule[2pt]
        \textbf{Simulator} & \bfseries\makecell[c]{RGB \\Camera} & \bfseries\makecell[c]{Depth \\Camera} & \bfseries\makecell[c]{Event \\Camera} & \textbf{LiDAR} & \textbf{Radar} \\ \midrule
        AirSim & $\surd$ & $\surd$ & - & - & - \\
        \rowcolor{whitesmoke} SVL & $\surd$ & $\surd$ & - & $\surd$ & $\surd$ \\
        Vista & $\surd$ & - & $\surd$ & $\surd$ & - \\
        \rowcolor{whitesmoke} CARLA & $\surd$ & $\surd$ & $\surd$ & $\surd$ & $\surd$ \\ \bottomrule[2pt]
    \end{tabular}  
    \quad
    \caption{A comparison of the ground truth annotations that actively maintained open-source simulators can synthesize.}
    \label{simulator-ground-truth}
    \begin{tabular}{l | c c c} \toprule[2pt]
        \textbf{Simulator} & \bfseries\makecell[c]{Bounding\\Box} & \bfseries\makecell[c]{Semantic\\Segmentation} & \textbf{Depth Map} \\ \midrule
        AirSim & - & $\surd$ & $\surd$ \\
        \rowcolor{whitesmoke} SVL & $\surd$ & $\surd$ & $\surd$ \\
        Vista & - & - & - \\
        \rowcolor{whitesmoke} CARLA & $\surd$ & $\surd$ & $\surd$ \\ \bottomrule[2pt]
    \end{tabular}
    \quad
    \caption{A comparison of the adverse weathers that actively maintained open-source simulators can synthesize.}
    \label{simulator-weather}
    \begin{tabular}{l | c c c c c} \toprule[2pt]
    \textbf{Simulator} & \textbf{Night} & \textbf{Cloudy} & \textbf{Rainy} & \textbf{Foggy} & \textbf{Snowy} \\ \midrule
    AirSim & - & - & - & - & - \\
    \rowcolor{whitesmoke} SVL & $\surd$ & - & $\surd$ & $\surd$ & - \\
    Vista & $\surd$ & - & $\surd$ & - & - \\
    \rowcolor{whitesmoke} CARLA & $\surd$ & $\surd$ & $\surd$ & $\surd$ & - \\ \bottomrule[2pt]
    \end{tabular}
\end{table}

As is shown in Table \ref{simulator-overview}, comprehensive simulators usually cover the functionalities of sensory data simulators. The key difference lies in the absence of quality vehicle dynamics simulation in sensory data simulators, making them unsuitable in end-to-end driving tasks \cite{muller2018sim4cv, paralleldomain, rong2020lgsvl}.

Notably, most sensory data simulators are no longer actively maintained. This trend may be attributed to the inherent technical challenges and costs of developing high-fidelity sensory data synthesis. As indicated in Table \ref{simulator-weather}, the creation of photo-realistic snowy scenarios remains a considerable challenge, and progress in generating high-fidelity adverse weather conditions is relatively slow \cite{von2019simulating, tremblay2021rain, hahner2022lidar}. Given this situation, researchers engaged in ADS tasks like object detection, semantic segmentation, instance segmentation, depth prediction, and trajectory tracking are advised to avoid complete reliance on simulator-generated data for training and testing purposes. Among the available simulators, CARLA stands out as the most suitable option for synthesizing data \cite{dosovitskiy2017carla}.

\subsection{Driving Policy Simulator}

Driving policy simulators are software that offers executable traffic scenarios for the development and evaluation of driving policy. The common functionalities of a driving policy simulator may include:

\begin{itemize}
    \item Traffic scenario running: These simulators enable the execution and visualization of driving policy models within specific traffic scenarios.
    \item Traffic status checking: Typically, driving policy simulators feature a traffic status checking capability, facilitating the validation of driving policy models.
    \item Real-world trajectory data import: Some driving policy simulator offers an interface for importing real-world trajectory data to enhance the fidelity of traffic scenarios.
    \item Traffic scenario customization: This function allows users to customize road layouts and vehicle behavior models to improve the diversity of traffic scenarios.
\end{itemize}

Table \ref{simulator-policy} overviews the most concerned features in actively maintained open-source simulators. \textit{Realistic perception} denotes simulators capable of producing photo-realistic RGB image output; otherwise, they provide a BEV semantic image. \textit{Custom map} represents that the simulator supports users to import maps in some popular data format like OpenStreetMap (OSM) and OpenDRIVE \cite{haklay2008openstreetmap, dupuis2010opendrive}. It is essential to note that most driving policy simulators lack a built-in scenario editor. If the users need to customize their traffic scenario, a third-party scenario editor is necessary. \textit{Trajectory importing} indicates whether the simulator is able to import trajectory data from at least one of the popular open trajectory datasets: HighD and its successors, INTERACTION, Nuplan, and Waymo Open Motion Dataset \cite{krajewski2018highd, bock2020ind, krajewski2020round, bock2021highly, xu2022drone, zhan2019interaction, caesar2021nuplan, ettinger2021large}. \textit{Multi-agent} shows whether the simulator supports multi-agent driving policy models. The number of available driving policy simulators is numerous. Readers can refer to the tables and select the most suitable ones based on their requirements.

\begin{table}[htb]
    \caption{A comparison of the actively maintained open-source driving policy simulators.}
    \label{simulator-policy}
    \centering
    \begin{tabular}{l | c c c c} \toprule[2pt]
        \textbf{Simulator} & \bfseries\makecell[c]{Realistic\\Perception} & \bfseries\makecell[c]{Custom\\Map} & \bfseries\makecell[c]{Trajectory\\Importing} & \bfseries\makecell[c]{Multi-\\agent} \\ \midrule[1pt]
        VDrift & $\surd$ & - & - & - \\
        \rowcolor{whitesmoke} CarRacing & - & - & - & - \\
        CommonRoad & - & $\surd$ & $\surd$ & $\surd$ \\
        \rowcolor{whitesmoke} highway-env & - & - & - & $\surd$ \\
        SMARTS & - & $\surd$ & - & $\surd$ \\
        \rowcolor{whitesmoke} L2R & $\surd$ & $\surd$ & - & - \\
        MetaDrive & $\surd$ & $\surd$ & $\surd$ & $\surd$ \\
        \rowcolor{whitesmoke} NuPlan & - & $\surd$ & $\surd$ & - \\
        InterSim & - & $\surd$ & $\surd$ & $\surd$ \\
        \rowcolor{whitesmoke} TBSim & - & $\surd$ & $\surd$ & $\surd$ \\
        Waymax & - & $\surd$ & $\surd$ & $\surd$ \\
        \bottomrule[2pt]
    \end{tabular}
\end{table}

\subsection{Vehicle Dynamics Simulator}

Vehicle dynamics simulators are software that can replicate the dynamic behavior of vehicles based on physics principles. These simulators consider both internal factors, such as the vehicle's powertrain, suspension, and aerodynamics, and external conditions, including road friction, wind resistance, and slope. By considering these variables, the simulator accurately predicts the dynamic characteristics of vehicles, like acceleration, braking, and steering response to different control commands.

Due to the well-established theories in vehicle dynamics, high-quality vehicle dynamics simulators have existed for a considerable period. However, owing to their reliance on real-world experimental data, the most reliable vehicle dynamics simulators are typically developed for commercial use in close collaboration with car manufacturers \cite{carsim, carmaker}.

Vehicle dynamics simulators find application in optimizing vehicle design and testing vehicle control algorithms. Among open-source options, Gazebo stands out as a reliable choice due to its rich experience in physics simulation in the robotics domain and its active community support \cite{michel2004cyberbotics}. Additionally, for academic institutions collaborating with MathWorks, Matlab proves to be a considerable choice for its integration capabilities with various simulators \cite{matlab}.

\subsection{Comprehensive Simulator}

\begin{figure*}[htb]
    \centering
    \includegraphics[width=\linewidth]{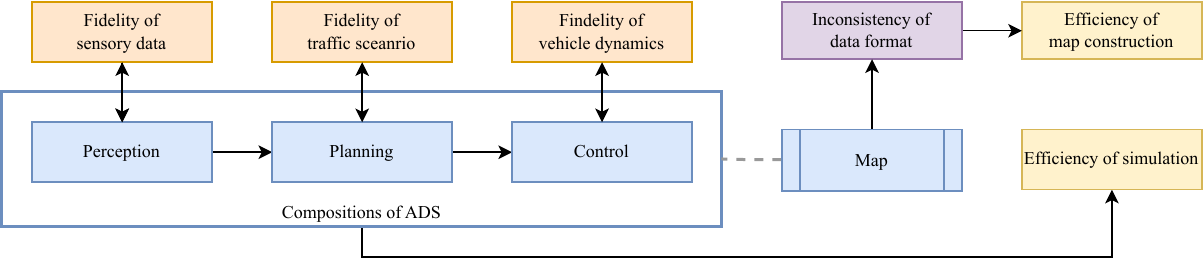}
    \caption{The critical issues of simulators and their relationship to the compositions of ADS.}
    \label{simulator-issues}
\end{figure*}

A comprehensive simulator is the integration of the functionalities discussed earlier. A comprehensive simulator can support multiple tasks across the perception, planning, and control parts of the ADS (see Table \ref{simulator-overview}). Figure \ref{simulator-category} illustrates how a comprehensive simulator performs Software-In-the-Loop simulation for various ADS tasks. 

The development of a reliable comprehensive simulator necessitates proficiency across multidisciplinary technologies. The demand for the ability to generate high-fidelity data for every task calls for a substantial commitment of human resources and financial support to the development process. Consequently, most of the qualified comprehensive simulators are for commercial use. The Virtual Test Drive (VTD) is a widely acclaimed comprehensive simulator offering highly integrated functionalities that satisfy diverse needs in developing an autonomous driving system \cite{vtd}. A similar product, SCAnER, presents comparable attributes, yet its popularity remains relatively limited due to a lack of promotion \cite{scaner}. The only open-source comprehensive simulator confirmed to be under maintenance is CARLA \cite{dosovitskiy2017carla}. At present, there is no substitute for CARLA in some specific tasks among the open-source simulators.

\section{Critical Issues and Improve Suggestions}
\label{issues}

This section will address the unresolved critical issues within simulators. The challenges faced by simulators can be summarized as three key aspects: fidelity, efficiency, and data format inconsistency. Fig. \ref{simulator-issues} illustrates the relationship between these critical issues and the modules of the ADS, emphasizing their widespread impact on the training and testing processes across various tasks within the ADS framework. 

The following subsections will explore the critical issues of the simulators, providing in-depth descriptions and justifications for their significance. Furthermore, the subsections will propose suggestions for improvement by drawing insights from existing literature.

\subsection{Fidelity of Sensory Data}
\label{issue-sensor-fidelity}

\subsubsection{Description}

Optical sensors, including cameras, LiDAR, and radar, play a crucial role in interpreting the dynamic traffic environment. Unfortunately, these sensors are vulnerable to adverse light and weather conditions \cite{vargas2021overview}. Factors like illumination, visibility, and humidity fluctuate in different weather, introducing noise into sensor data, so the detection range and precision are compromised \cite{goodin2019predicting}. Adverse weather conditions also alter the reflectance of roads, further perturbing the efficacy of perception algorithms \cite{zang2019impact}. Hence, enhancing the robustness of perception algorithms in diverse light and weather conditions has been a longstanding objective.

The process of collecting sensory data from reality is costly and time-consuming. Given the unpredictable nature of weather, it is common that the sensory data under demanded conditions is not accessible. In contrast, simulators offer precise control over weather conditions through adjustable parameters, eliminating the need for extended waits to encounter specific weather scenarios \cite{kalra2016driving}. Moreover, the labor-intensive task of manually annotating ground truth data is obviated. These inherent advantages suggest that if simulators can proficiently generate high-fidelity data in adverse weather conditions, they have the potential to substitute real-world data and promote relevant research \cite{li2022toward, goodin2019predicting}.

\begin{figure*}[htb]
    \centering
    \subfloat{
        \includegraphics[width=0.23\textwidth]{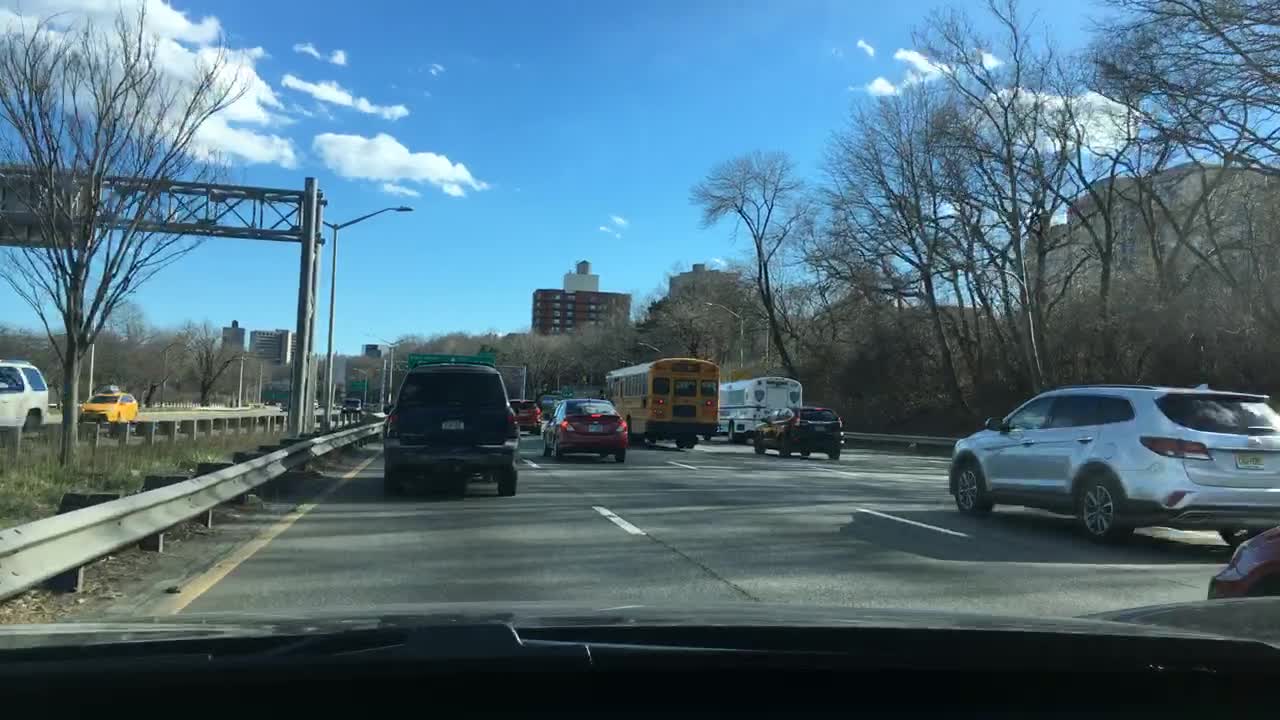}
    }
    \subfloat{
        \includegraphics[width=0.23\textwidth]{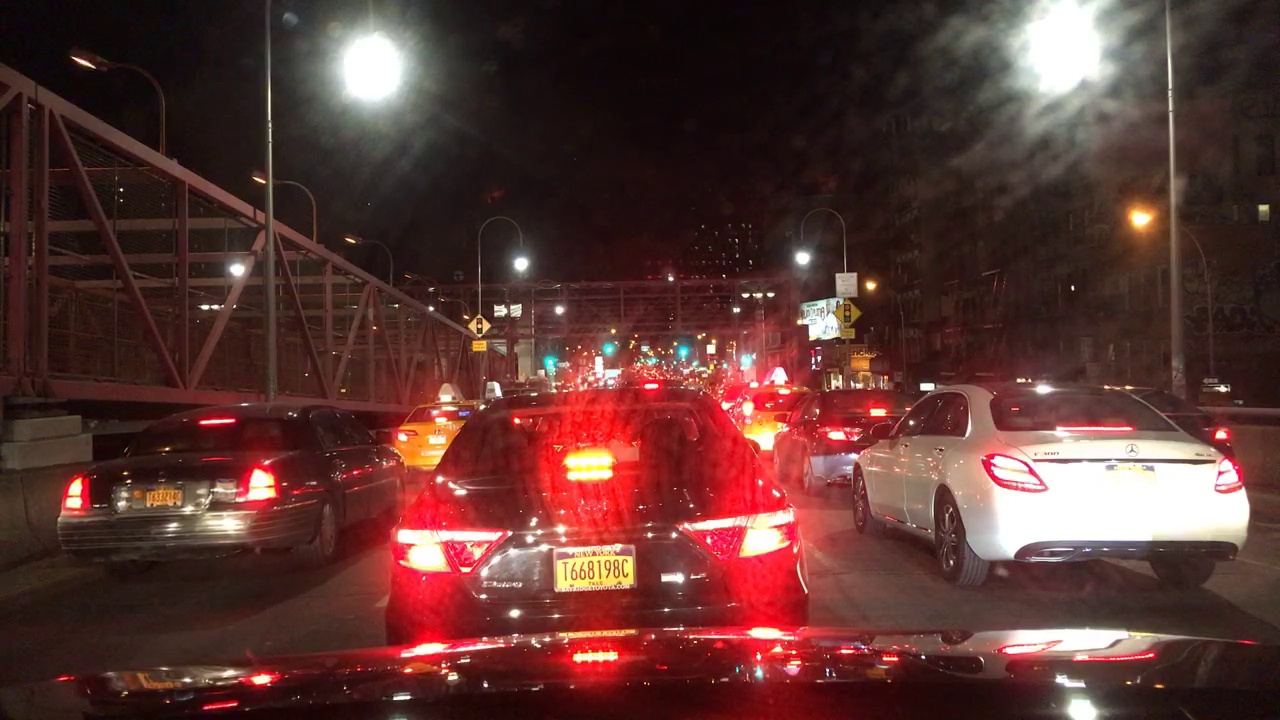}
    }
    \subfloat{
        \includegraphics[width=0.23\textwidth]{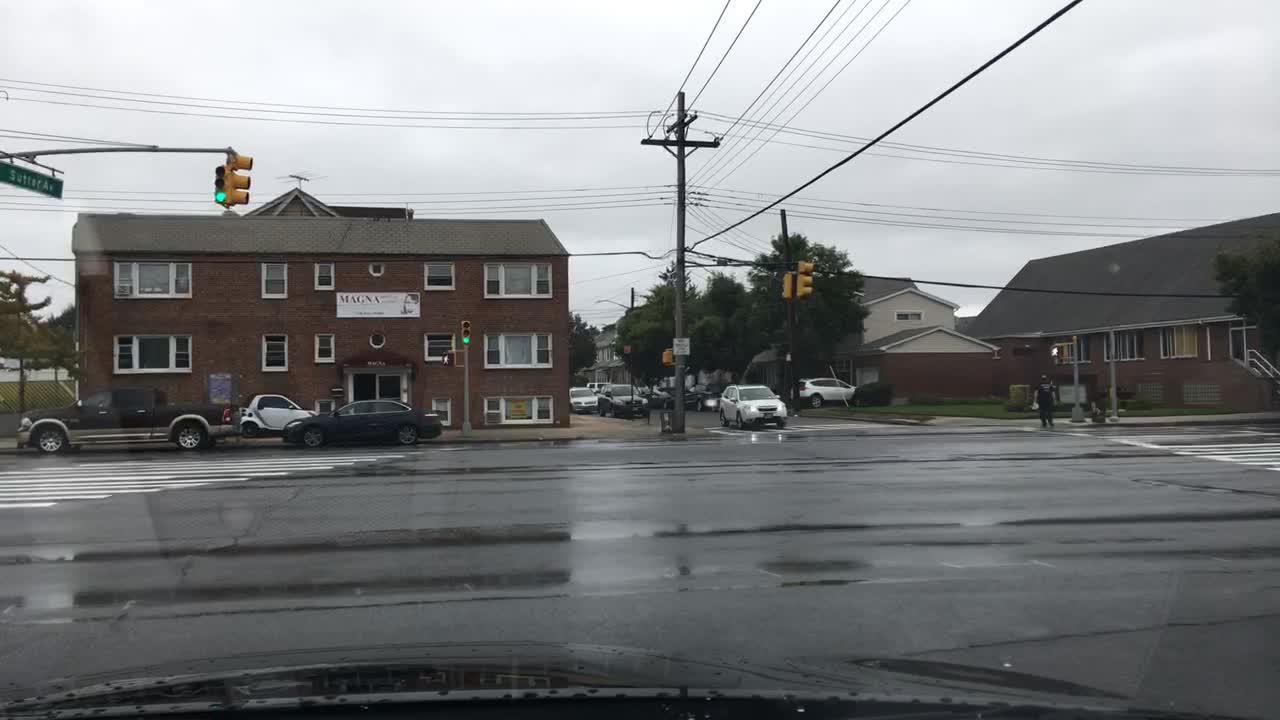}
    }
    \subfloat{
        \includegraphics[width=0.23\textwidth]{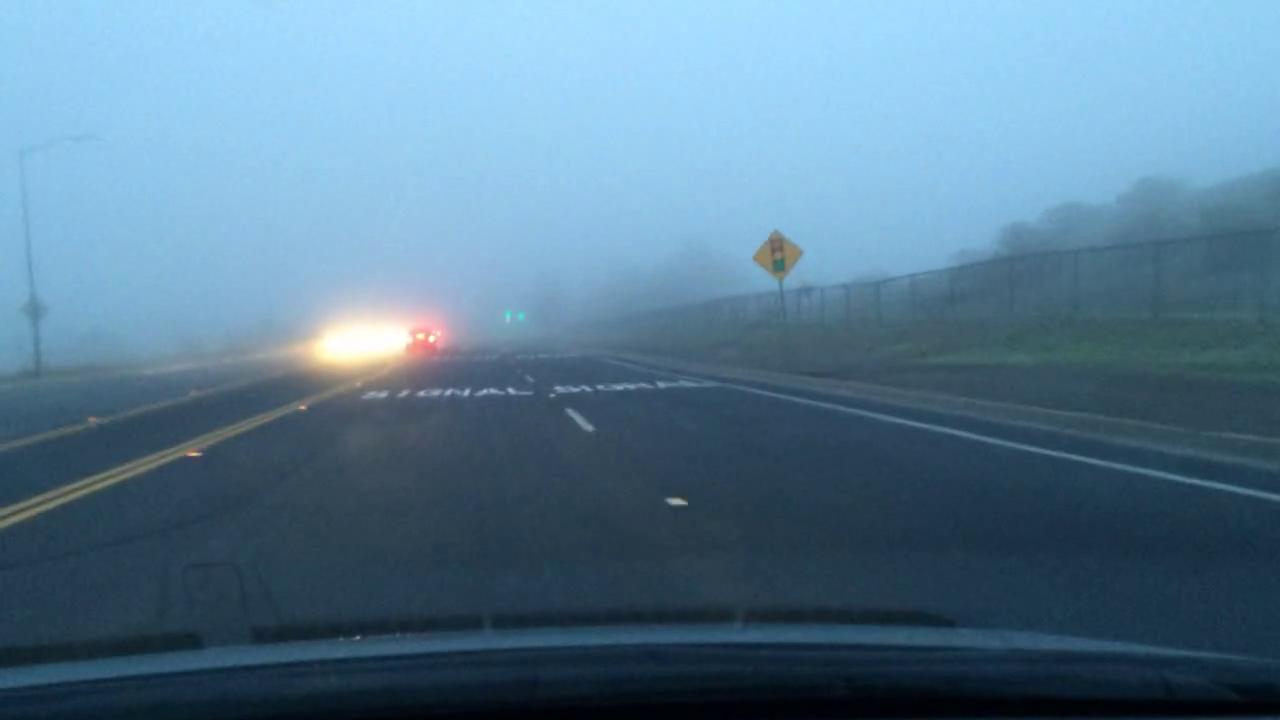}
    }
    \quad
    \subfloat{
        \includegraphics[width=0.23\textwidth]{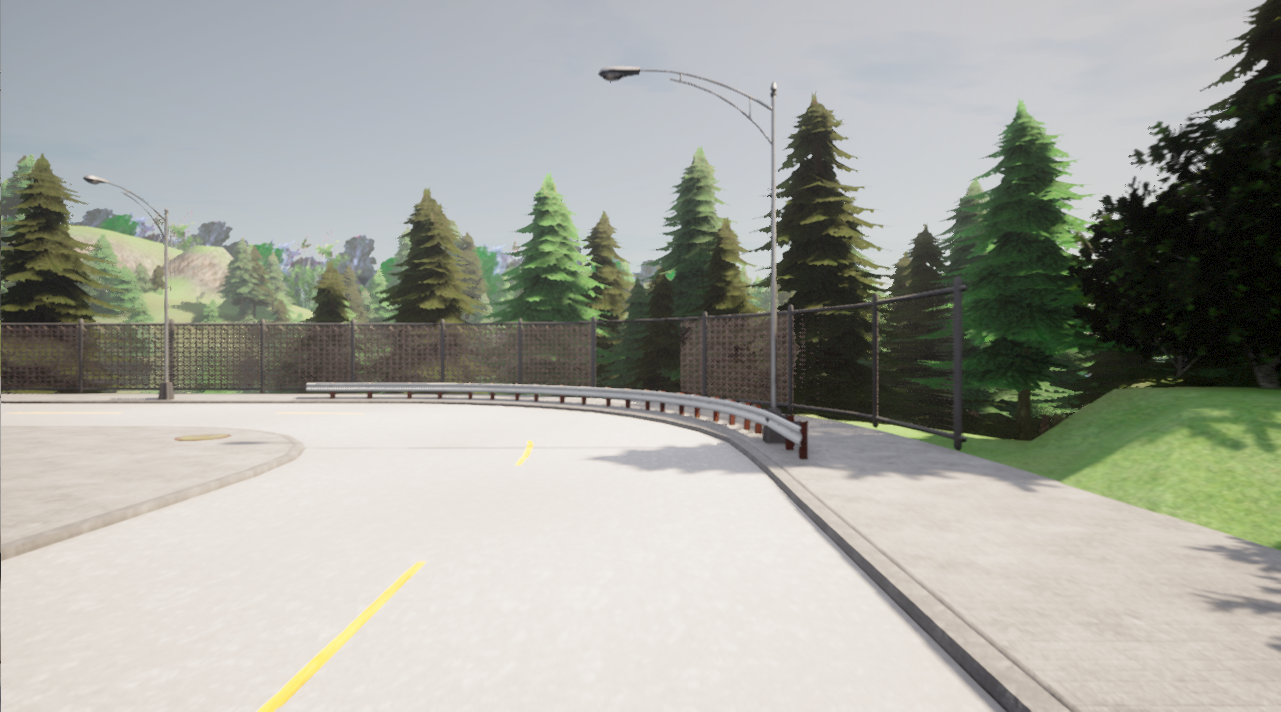}
    }
    \subfloat{
        \includegraphics[width=0.23\textwidth]{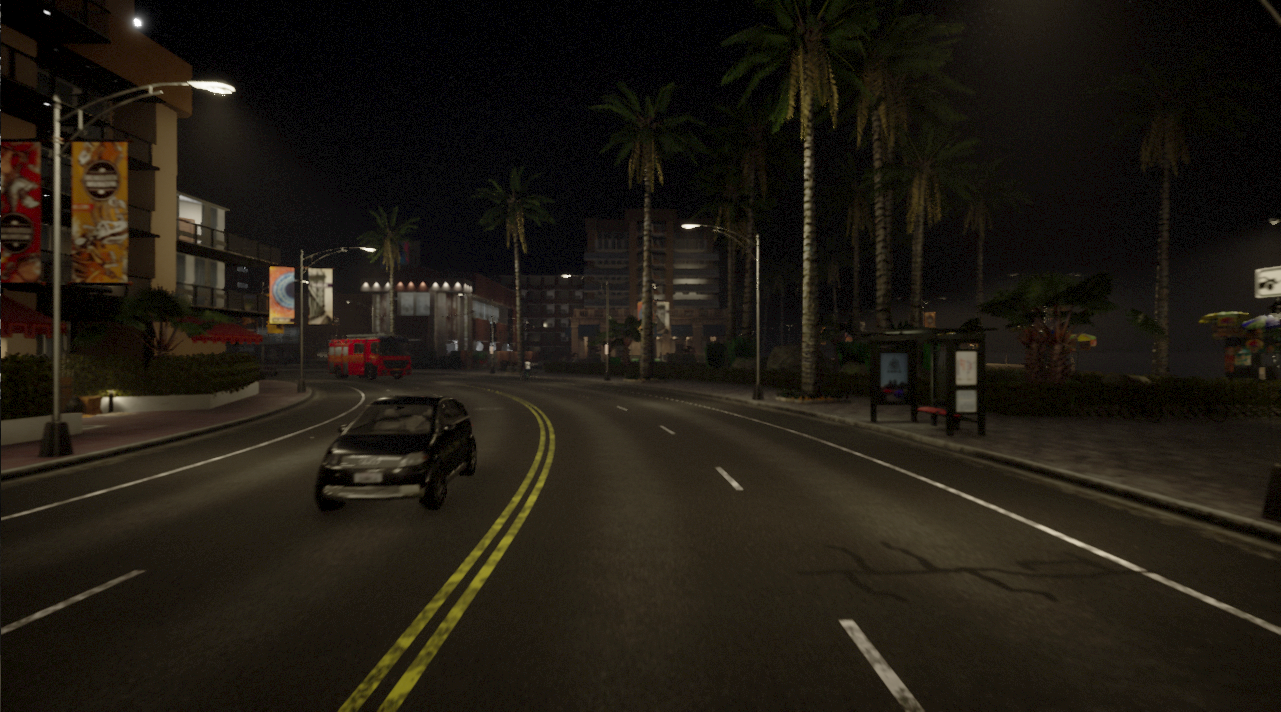}
    }
    \subfloat{
        \includegraphics[width=0.23\textwidth]{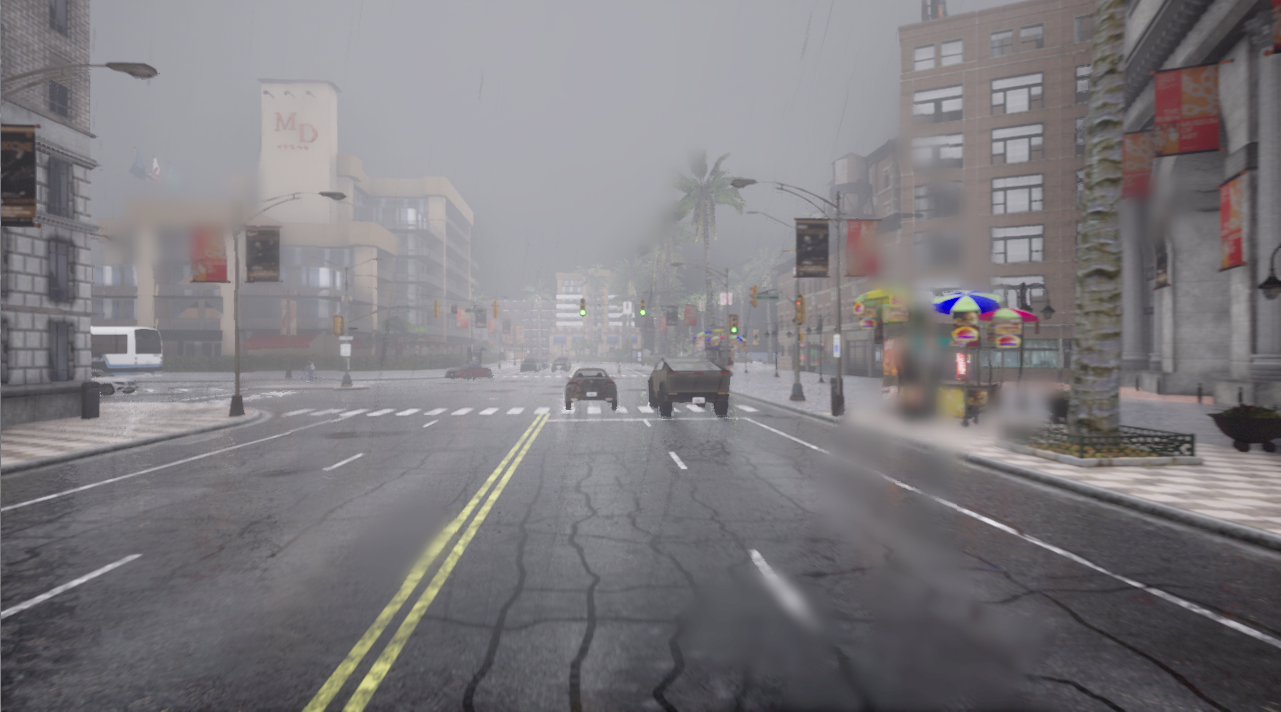}
    }
    \subfloat{
        \includegraphics[width=0.23\textwidth]{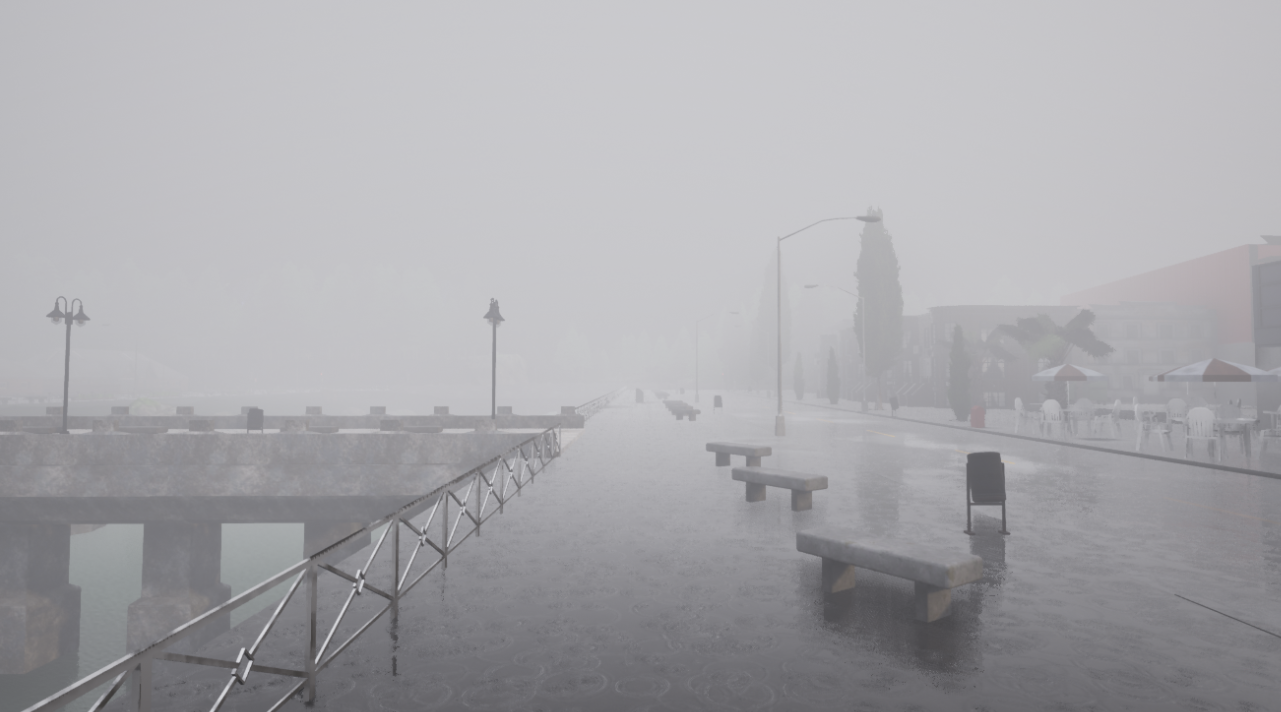}
    }
    \caption{A comparison of real-world and simulated RGB images in day, night, rainy, and foggy weather. The first row are from BDD100K dataset and the second row is from CARLA \cite{yu2020bdd100k, dosovitskiy2017carla}}
    \label{issue-weather-comparison}
\end{figure*}

In the context of synthesized sensory data, \textit{fidelity} describes how faithful the synthesized data is to reality. As indicated in Table \ref{simulator-sensor}, only a few simulators can generate sensory data under complex conditions. Due to technology limitations, the simulators still struggle to present the functionality of adverse weather simulation, with the quality of implementation often overlooked. Fig. \ref{issue-weather-comparison} shows the fidelity gap between the RGB images obtained from the leading simulator CARLA and the real-world dataset BDD100K \cite{dosovitskiy2017carla, yu2020bdd100k}. This gap explains why the existing simulators may not be qualified for robust perception model training and testing.

The loss of fidelity of sensory data is closely related to the architecture of simulators. Since these simulators commonly leverage game engines for development, a prevalent practice is to utilize default particle systems within these engines to simulate adverse weather conditions \cite{rong2020lgsvl, shah2018airsim, dosovitskiy2017carla}. In a particle system, rain and snow are treated as particles with physics properties. During the simulation process, the emphasis is on the movement trajectory of particles influenced by wind speed and gravity, but less attention is paid to the texture pattern \cite{dunkerley2008rain, von2019simulating, tremblay2021rain}. Furthermore, existing weather simulations often only impact rendering, meaning they can only generate image data for adverse weather without adding corresponding noise interference to LiDAR, radar, etc.

\subsubsection{Improvement suggestions}

To improve the fidelity of sensory data simulation, an intuitive idea is to study the realistic noise pattern. In most synthesis processes, the distribution pattern of noise is first extracted, then a mask is generated based on the distribution, and the original data merges with the mask through post-processing. One of the earliest practices involved modeling the noise distribution pattern with particle systems and physical properties \cite{starik2003simulation, wang2006real, von2019simulating}, which was initially applied to image generation and extended to enhance LiDAR data \cite{hahner2022lidar}.

In recent years, there has been an increase in learning-based data synthesis methods. This evolution began by employing Deep Neural Networks (DNN) to introduce variations to filters, addressing rain, fog, and camera-blurring situations \cite{tian2018deeptest, sakaridis2018semantic}. K-Radar proposed a Neural Network (NN)-based approach to generate adverse weather data specifically for Radar applications \cite{paek2022k}. As the quest for improved fidelity and complexity intensified, Adversarial Generative Networks (GAN) were introduced to the synthesis task. GANs enabled the generation of artificial instances of snowing occasions \cite{zhang2018deeproad, mușat2021multi}. The study by \cite{yang2020surfelgan} underscored the potential of GANs in generating RGB images under various lighting conditions, including day, night, and dawn. Tremblay et al. sought to combine physics rules with GAN techniques, allowing for control over precipitation in synthesized images \cite{tremblay2021rain}. This synthesis approach successfully maintains visual realism while adhering to fundamental physical laws. ClimateNerf serves as the first attempt to apply NeRF to simulate adverse weather conditions, which achieves realistic 3D rendering results, particularly in simulating flood and snow \cite{li2022climatenerf}.

Sensory data simulation algorithms are often available as open-source projects online \cite{paek2022k, li2022toward, li2022climatenerf}. However, these algorithms exhibit variations in environment configurations and hardware requirements, presenting challenges for users seeking quick adoption. If the simulators could establish a standardized interface that defines the input and output of algorithms, it would simplify the process for users and foster a more seamless incorporation of diverse synthesis models into the simulation framework.

\subsection{Fidelity of Traffic Scenario}
\label{issue-scenario-fidelity}

\subsubsection{Description}

The main components of a traffic scenario are the geometry map, traffic rubrics, and traffic participants. Concerning the static parts, the geometry map, and traffic rubrics, fidelity is not the primary challenge since manual annotation can effectively capture information, including temporary changes. When handling this static data, the efficiency issue is a topic to explore in subsection \ref{issue-scenario-efficiency}. The dynamic part of the traffic scenario faces the fidelity problem. Intricate movements and interactions among traffic participants resist explicit depiction. Defined by inherent randomness and individual variations, the dynamic behavior of traffic participants becomes a principal source of complexity of traffic scenarios. 

The traffic participants refer to the dynamic objects in a traffic scenario, such as vehicles, cyclists, and pedestrians. Ideally, these participants should be able to perceive their surroundings, comprehend the actions of other participants, and navigate within the scenario \cite{ghodsi2021generating}. 

A significant limitation of the ADS simulators lies in the lack of \textit{fidelity} in traffic participants. Currently, two mainstream methods exist to generate a traffic scenario in simulators. One approach is replicating actual trajectories, resetting the environment when a tested agent violates a rule or experiences a collision \cite{caesar2021nuplan}. However, this method assumes that the replayed trajectory is an optimal solution, potentially limiting the planners to human-level behavior. Another way is to rely on rule-based behavioral models, which exhibit simplistic behavior patterns \cite{kesting2007general, dosovitskiy2017carla}. These models are constrained to actions like maintaining a specific distance and stopping, lacking the flexibility to circumvent obstacles or accelerate through scenes \cite{krauss1998microscopic, erdmann2015sumo, panwai2005comparative}. The notable disparity between the behavior of simulated traffic participants and their real-world counterparts undermines the credibility of simulation experiments in planning tasks \cite{siebke2022traffic}.

\subsubsection{Improvement suggestions}

The key to enhancing the fidelity of dynamic object movements is to refine behavior modeling with realistic random events \cite{schwarting2018planning, bhattacharyya2021hybrid}. Lee et al. proposed to treat the modeling process as a trajectory prediction task, leveraging Graphical Neural Networks (GNN) to learn patterns from real-world datasets \cite{lee2019joint}. \cite{kumar2021interaction} followed this idea and implemented TrafficGraphNet, which is capable of temporal interactions in a sparse traffic scenario. Similarly, VectorNet and InterSim employ deeper and customized NN to extract motion patterns from datasets \cite{gao2020vectornet, sun2022intersim}.

To alleviate a dependency on an initial dataset, researchers have explored the concept of adversarial to interaction generation. RouteGAN applied GAN to produce styled behavior for individual vehicles \cite{yin2021diverse}. Soon, the researchers realized that inverse Reinforcement Learning (RL) structure could adopt a similar idea while better fitting in the time-sequential decision-making \cite{shiroshita2020behaviorally}. A deep deterministic policy gradient was tried to introduce complexity in lane-changing scenarios \cite{chen2021adversarial}. Trajgen pointed out that the combination of trajectory prediction guarantees fidelity, while the RL method adds flexibility to achieve a realistic and diverse trajectory output \cite{zhang2022trajgen}.

Generating interactive traffic participants remains a prevalent research topic \cite{ding2023survey}. The researchers are developing various architectures to elevate the sophistication and realism of the generated behaviors. In this context, empowering open-source simulators to provide a user-friendly interface for customizing traffic participants could prove beneficial. By equipping simulators with a reliable traffic event detector and a framework for tailoring behavior models, the research community can facilitate realizing more interactive traffic participant controllers through collaborative contributions.

\subsection{Fidelity of Vehicle Dynamics}
\label{issue-dynamics-fidelity}

\subsubsection{Description}

A reliable vehicle dynamics model is critical to guarantee that the planning result of the ADS is executed as expected. The fidelity of vehicle dynamics is a particularly prevalent problem in open-source simulators, while it has been well resolved in commercial vehicle dynamics simulators. This discrepancy may arise from the demanding nature of constructing an accurate vehicle dynamics model, which necessitates substantial measurement data, particularly under extreme conditions such as sharp turns and sudden stops at high speeds combined with diverse weather and road conditions \cite{spielberg2023learning}. Undertaking this endeavor is costly and raises safety concerns, making it challenging to accomplish without collaborative support from vehicle manufacturers.

Table \ref{simulator-dynamics} investigates the implementation status of vehicle physics models in actively maintained open-source simulators. Half of the simulators integrate a third-party physics engine \cite{smith2005open, sanders2016introduction, parberry2017introduction, coumans2021pybullet, goslin2004panda3d}. The rest choose to implement a dynamics/kinematics bicycle model independently. Most of these physics models are faithful under conditions where velocity, acceleration, and steering angle values are small \cite{kong2015kinematic}. This suggests a limitation in existing open-source simulators, impeding their ability to test the control module of ADS. This constraint, in turn, hinders the effective implementation of planning algorithms on actual vehicles.

\begin{table}[htb]
    \caption{A comparison of the physics model/dynamics model of the actively maintained open-source simulators.}
    \label{simulator-dynamics}
    \centering
    \begin{tabular}{l | c c} \toprule[2pt]
        \textbf{Simulator} & \textbf{Physics Engine} & \textbf{Bicycle Model} \\ \midrule[1pt]
        SUMO & - & - \\
        \rowcolor{whitesmoke}  Webots & Open Dynamics Engine \cite{smith2005open} & - \\
        VDrift & Custom & - \\
        \rowcolor{whitesmoke}  CarRacing & Box2D \cite{parberry2017introduction} & - \\
        CommonRoad & - & $\surd$ \\
        \rowcolor{whitesmoke}  highway-env & - & $\surd$ \\
        SMARTS & PyBullet \cite{coumans2021pybullet} & $\surd$ \\
        \rowcolor{whitesmoke}  L2R & Unreal Engine 4 \cite{sanders2016introduction} & - \\
        MetaDrive & Panda3D \cite{goslin2004panda3d} & - \\
        \rowcolor{whitesmoke}  NuPlan & - & $\surd$ \\
        InterSim & - & $\surd$ \\
        \rowcolor{whitesmoke}  TBSim & - & $\surd$ \\
        Waymax & - & $\surd$ \\
        \rowcolor{whitesmoke}  CARLA & Unreal Engine 4 & - \\
        \bottomrule[2pt]
    \end{tabular}
\end{table}

\subsubsection{Improvement suggestions}

Given that creating a realistic vehicle dynamics model relies on real-world data fraught with cost and safety concerns, it is hard for open-source simulators to achieve high-fidelity representations. A possible solution lies in establishing a platform where researchers and institutions can contribute locally collected real-vehicle dynamics data. A similar data-sharing endeavor has successfully occurred in structured maps and vehicle 3D models, which suggests the feasibility of such an approach \cite{haklay2008openstreetmap, koenig2004design}.

\subsection{Inconsistency of Data Format}
\label{issue-format}

\subsubsection{Description}

Most open-source simulators employ a self-defined data format to record maps and scenarios. Suppose the publicly available data format is defined as one that has an open document that describes all labels and annotations clearly. In that case, it becomes evident from Table \ref{simulator-data-format} that actively maintained open-source simulators provided limited support for such data formats \cite{behrisch2011sumo, haklay2008openstreetmap, dupuis2010opendrive, poggenhans2018lanelet2, openscenario, fremont2019scenic}. The inconsistency in map and scenario formats is problematic, as it necessitates redundant efforts in manually constructing traffic scenarios and maps, even when the data is already accessible. The absence of an official data conversion interface also introduces errors during the data transition process.

\begin{table*}[htb]
    \centering
    \caption{A comparison of the public map and scenario data formats that the actively maintained open-source simulators support.}
    \label{simulator-data-format}
    \begin{tabular}{l | c c c c | c c} \toprule[2pt]
        \multirow{2}*{\textbf{Simulator}} & \multicolumn{4}{c}{\textbf{Map Data Format}} & \multicolumn{2}{c}{\textbf{Scenario Data Format}} \\ \cmidrule{2-7}
        & SUMO Road Network \cite{behrisch2011sumo} & OpenStreetMap \cite{haklay2008openstreetmap} & OpenDRIVE \cite{dupuis2010opendrive} & Lanelet2 \cite{poggenhans2018lanelet2} & OpenSCENARIO \cite{openscenario} & Scenic \cite{fremont2019scenic} \\ \midrule
        SUMO & $\surd$ & $\surd$ & $\surd$ & - & - & - \\
        \rowcolor{whitesmoke}  Webots & $\surd$ & $\surd$ & - & - & - & - \\
        VDrift & - & - & - & - & - & - \\
        \rowcolor{whitesmoke}  CarRacing & - & - & - & - & - & - \\
        CommonRoad & $\surd$ & $\surd$ & $\surd$ & $\surd$ & $\surd$ & - \\
        \rowcolor{whitesmoke}  highway-env & - & - & - & - & - & - \\
        SMARTS & $\surd$ & - & - & - & - & - \\
        \rowcolor{whitesmoke}  L2R & - & - & - & - & - & - \\
        MetaDrive & - & - & - & - & - & - \\
        \rowcolor{whitesmoke}  NuPlan & - & - & - & - & - & - \\
        InterSim & - & - & - & - & - & - \\
        \rowcolor{whitesmoke}  TBSim & - & - & - & - & - & - \\
        Waymax & - & - & - & - & - & - \\
        \rowcolor{whitesmoke}  CARLA & - & - & $\surd$ & - & $\surd$ & $\surd$ \\ \bottomrule[2pt]
    \end{tabular}
\end{table*}

\subsubsection{Improvement suggestions}

The underlying problem behind the inconsistency of data format is that the existing data structures fail to fulfill the need to express the scenarios comprehensively and efficiently. The advantages and shortcomings of the map data format are as follows:

\begin{itemize}
    \item \textit{SUMO Road Network} is one of the earliest publicly available map data formats \cite{behrisch2011sumo}. It emphasizes the relationship between roads and junctions and provides rich annotation for traffic rule restrictions. However, its utility focuses on traffic flow simulation tasks, leading to a lack of detailed geometry information concerning roads and irregular static obstacles.
    \item \textit{OSM} originates from a free mapping project \cite{haklay2008openstreetmap}. Despite its inclusion of support for static obstacles and regions, the fundamental structure of OSM draws inspiration from road network representation. Consequently, it continues to employ a centerline to represent lanes, thereby falling short in high-resolution geometry descriptions of lanes.
    \item \textit{OpenDRIVE} has gained popularity due to its rich parameters and clear documentation \cite{dupuis2010opendrive}. OpenDRIVE excels in recording high-resolution map geometry but overlooks traffic control and road semantic information.
    \item \textit{Lanelet2} tempts to integrate the strong points of the data formats mentioned above together \cite{poggenhans2018lanelet2}. It accurately describes geometry structures and versatile, customizable semantic traffic contexts. However, it has not attracted as much support as OpenDRIVE, potentially because Lanelet2 lacks a corresponding scenario data format to articulate the dynamic objects.
\end{itemize}

For scenario data formats, \textit{OpenSCENARIO} provides a precise though intricate description of dynamic objects \cite{openscenario}. In this format, each traffic participant strictly runs on the assigned trajectory, ensuring a detailed representation of scenarios. In contrast, \textit{Scenic} emphasizes introducing randomness into scenarios \cite{fremont2019scenic}. It enables the generation of traffic scenarios by illustrating the probability distribution of behaviors.

Upon examining the current landscape of data formats, it becomes evident that the roadmap for enhancing map data formats involves combining lane-level representation accuracy with comprehensive traffic control and semantic information. Simultaneously, improvements in scenario data formats should accommodate both accurate scenario replay and flexible random movements. A seamless integration between map data and scenario representation demands further attention and dedication. Furthermore, developing an official implementation for a data parser is deemed necessary.

\subsection{Efficiency of Map Construction}
\label{issue-scenario-efficiency}

\subsubsection{Description}

\begin{figure*}
    \centering
    \subfloat[SUMO Netedit \cite{krajzewicz2005preparation}]{
        \includegraphics[height=4.6cm]{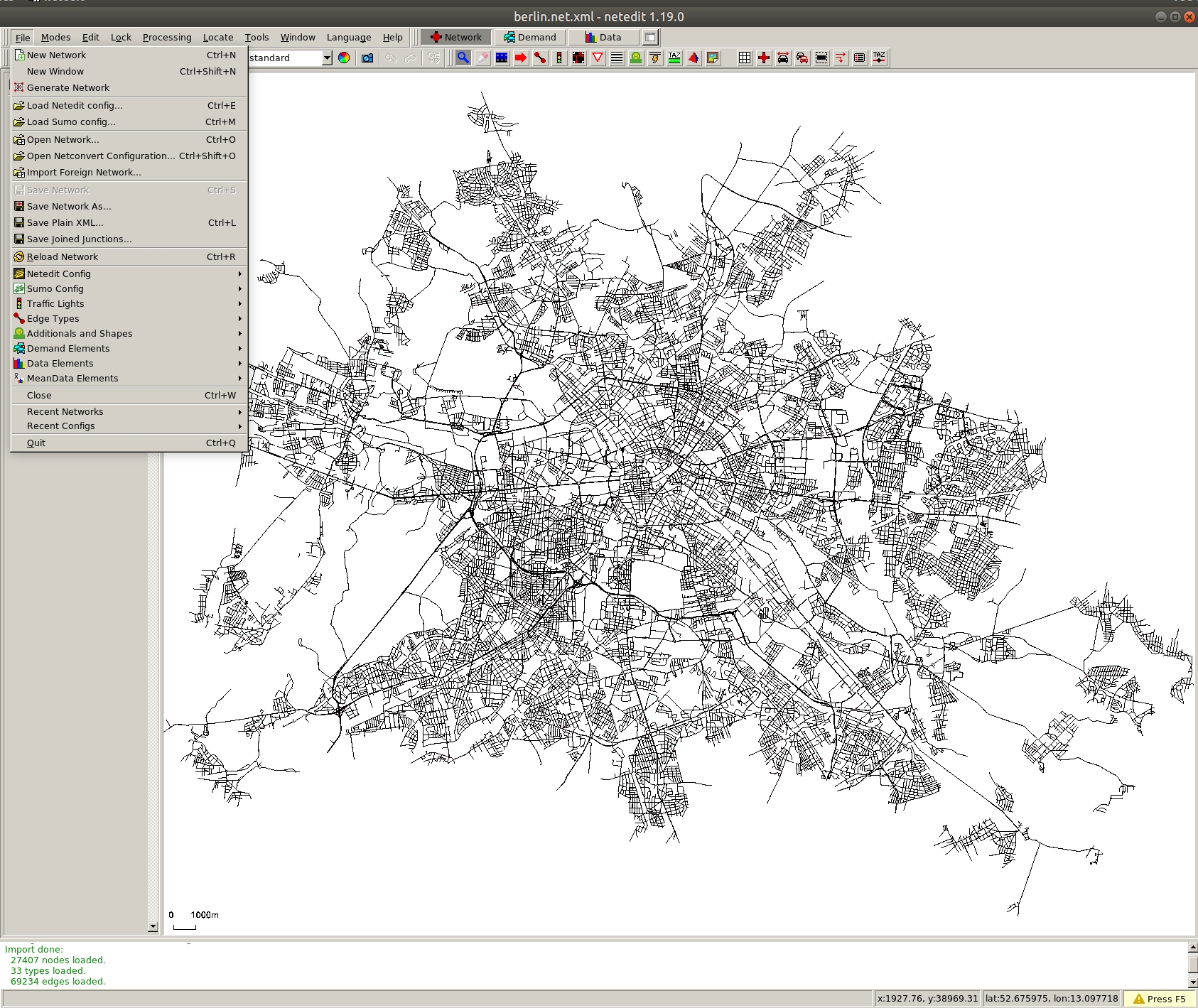}
    }
    \subfloat[CommonRoad Scenario Designer \cite{maierhofer2021commonroad}]{
        \includegraphics[height=4.6cm]{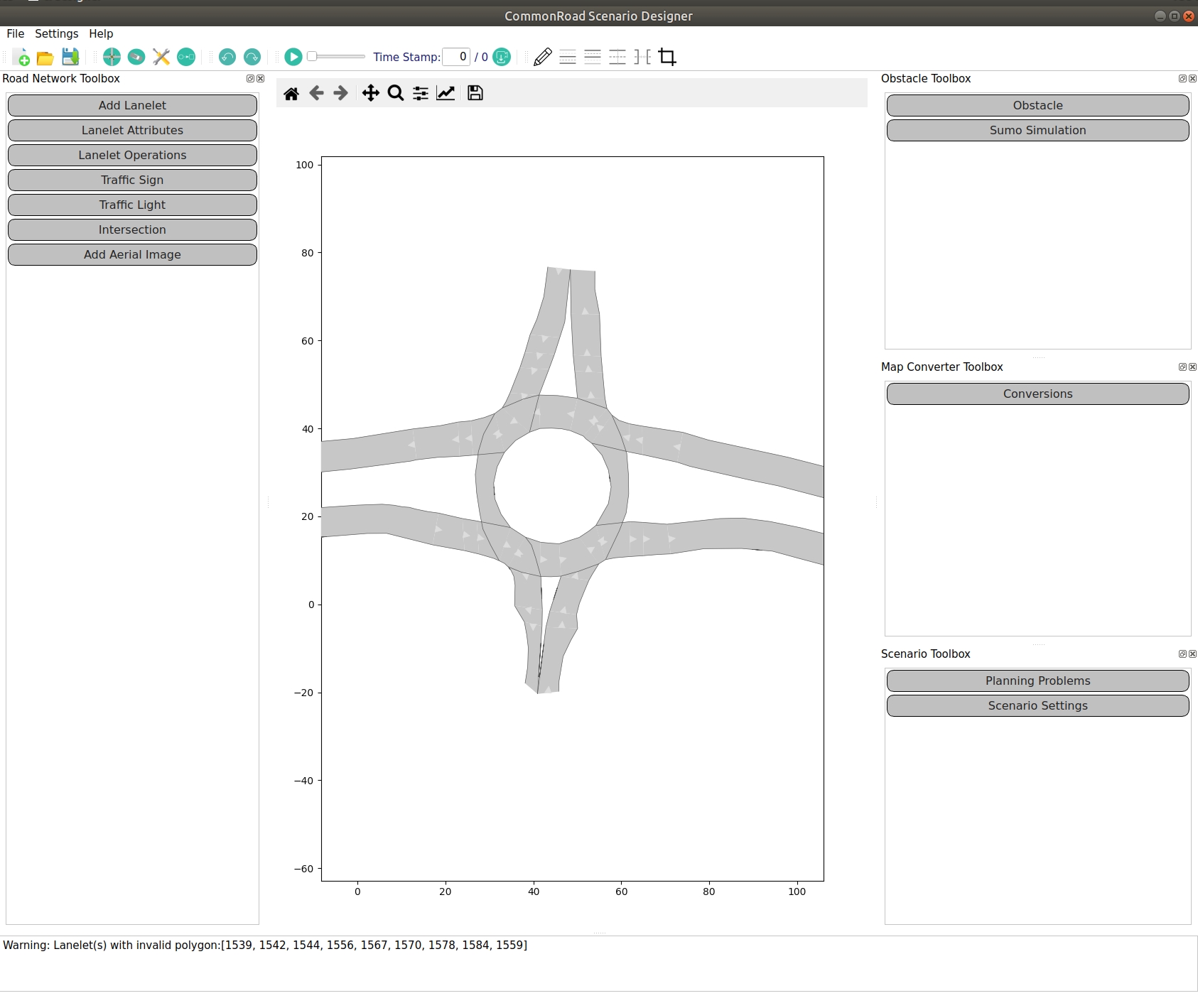}
    }
    \subfloat[RoadRunner \cite{roadrunner}]{
        \includegraphics[height=4.6cm]{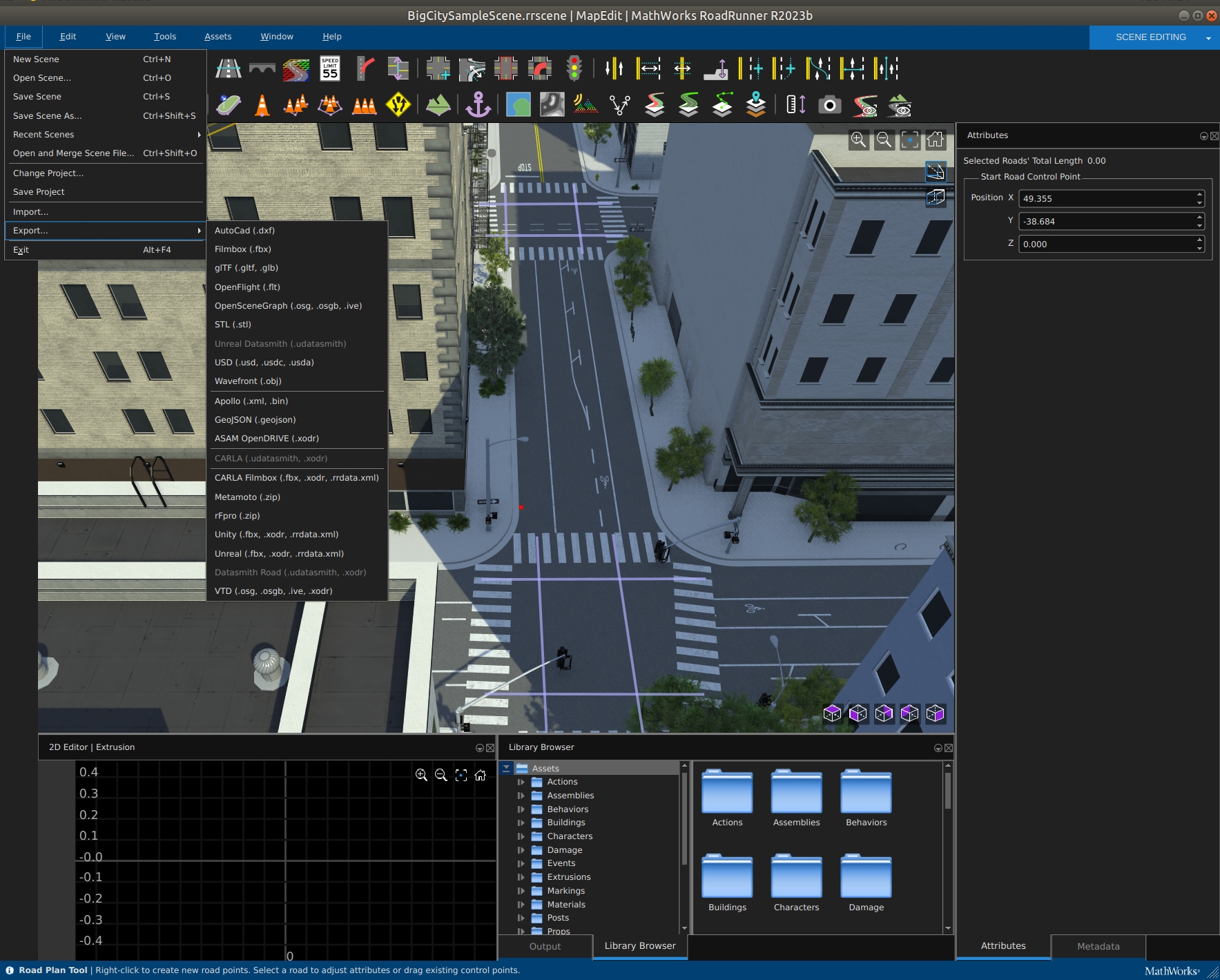}
    }
    \caption{The Graphical User Interface (GUI) of the scenario editors.}
    \label{issue-scenario-generation}
\end{figure*}

The simulators for autonomous driving are expected to have the ability to quickly rebuild or synthesize diverse traffic scenarios for training and testing purposes \cite{ding2023survey}. While the dynamic objects in traffic scenarios struggle with the fidelity issue (as discussed in Section \ref{issue-scenario-fidelity}), the static maps' construction also faces efficiency challenges. Fig. \ref{issue-scenario-generation} demonstrates the popular scenario editors \cite{krajzewicz2005preparation, maierhofer2021commonroad, roadrunner}. Currently, the preferred map reconstruction process requires manually extracting road structures, labeling road network relationships, and annotating traffic signals from LiDAR points and images \cite{huang2022city3d}. This approach heavily relies on handcraft and is inherently time-consuming.

Another challenge in efficiently creating high-quality maps is related to data formats. The common output of 3D map reconstruction, a 3D mesh model, often lacks detailed structural information about roads and buildings \cite{nan2017polyfit, pan2019deep}. Moreover, semantic annotations for elements like signs and signals are not inherently present within the 3D mesh model. Meanwhile, although 2D map reconstruction approaches provide valuable data on road structure, seamlessly integrating the outputs of 2D reconstruction with 3D map reconstruction poses challenges. This is primarily because the 2D map format typically uses polygons to depict geometric structures, while 3D models utilize triangular meshes \cite{poggenhans2018lanelet2, haklay2008openstreetmap, lee2019study}. Consequently, aligning the 2D road structure with the 3D mesh model becomes complex.

\subsubsection{Improvement suggestions}

Automated map reconstruction utilizes algorithms to process 2D pixels and 3D points and generate structured maps. Achieving partial or complete automated map construction can significantly reduce manual effort and improve the efficiency of building scenarios.

Automated map reconstruction is advancing along two distinct trajectories. One avenue is oriented towards 3D reconstruction, characterized by an emphasis on achieving high-level realism and details. The general idea is to extract features from 2D images and match them with 3D points and pose information. By establishing correlations between object models and textures, this method enables effective automatic map reconstruction \cite{fruh2004automated, snavely2006photo, agarwal2011building}. Traditional 3D map reconstruction methods use explicit priors to process large-scale raw data rapidly \cite{agarwal2011building, williams2021neural}. The new rising learning-based methods extract data-driven priors by neural networks to handle noise and data sparsity \cite{pan2019deep, boulch2022poco}. A recent breakthrough in this area is the NeRF algorithm. It learns the implicit features of space from images captured from different angles, representing 3D space as a continuous radiance field \cite{tancik2022block, huang2023neural}. NVIDIA has implemented 3D surface reconstruction and rendering based on the NeRF algorithm, demonstrating its great potential for automated 3D map reconstruction \cite{li2023neuralangelo}.

Another avenue focuses on reconstructing the road structure and traffic information, exhibiting a greater tolerance for deviation in the map's geometry details \cite{mondal2020real2sim}. These techniques prioritize comprehensively depicting the traffic environment, encompassing traffic elements such as the road network, traffic signals, and dynamic objects \cite{tan2021scenegen, ghodsi2021generating}. With the overarching structure and interplay dynamics of the traffic scenario, the 2D maps generated by these approaches become preferable for driving decision-making algorithms \cite{gao2020vectornet, tian2022real}.

These two categories of automated map reconstruction methods can reduce the need for manual intervention in constructing simulation scenarios for autonomous driving \cite{kar2019meta}. However, they are unsuitable for practical implementation at present. The 3D map construction techniques exhibit sensitivity to data distribution and susceptibility to noise interference \cite{tan2021scenegen}. The precision and speed of these methods fall short of the standard for industrial application \cite{tancik2022block}.

\subsection{Efficiency of Simulator}
\label{issue-efficiency}

\subsubsection{Description}

Training and testing data-driven ADS tasks are time-consuming due to the need for algorithms to cover a variety of scenarios, ensuring that challenging corner cases are addressed \cite{xu2017end}. The estimated 215 billion miles for validating an ADS highlights the necessity to apply a simulator to speed up this process \cite{kalra2016driving}. There are several common strategies to improve simulation efficiency:

\begin{itemize}
    \item \textit{Hardware acceleration} accelerates the simulation process by providing more powerful computational resources. An upgrade to the CPU, GPU, and memory devices could expedite the computational operations and strengthen the capability to handle data batches \cite{gulino2023waymax}.
    \item \textit{Computation optimization} is to make better utilization of the computational resource. It streamlines the simulation workflow by fine-tuning the algorithm, code structure, and cache management \cite{varga2023optimizing}.
    \item \textit{Parallelization} maximize the utility of the computational resource at a lower level. It enables simultaneous execution of multiple threads. This approach would significantly reduce the simulation time by distributing the workload across multiple cores or processors \cite{rousset2016survey}.
    \item \textit{Distributed computation} deploys large-scale computation resources over different machines. Running multiple simulators concurrently for training or testing individual models serves to minimize the overall running time, enhancing the efficiency of the process \cite{rousset2016survey}.
\end{itemize}

Additionally, \textit{headless simulation}, also known as off-screen simulation, may become an unavoidable consequence for achieving efficient simulation, as powerful hardware and distributed computation often depend on servers without display. 

A field investigation of the actively maintained open-source simulator shows that most of the simulators lack the necessary features for high-efficiency simulation. Table \ref{simulator-efficiency} provides insights into the minimal hardware requirements for installation disk space, the memory occupancy during tutorial running, and the GPU memory occupancy, which reflects how well the simulator architecture has been optimized. According to the official documentation, most simulators are now cross-platform compatible, but few of them support hardware acceleration, distributed computation, and headless simulation.

\begin{table*}[htb]
    \caption{A comparison of the hardware requirements and performance of the actively maintained open-source-source simulators.}
    \label{simulator-efficiency}
    \centering
    \begin{tabular}{l | c c c | c c c | c c c}
    \toprule[2pt]
        \multirow{2}*{\textbf{Simulator}} & \multicolumn{3}{c}{\textbf{Minimal Hardware Requirements}} & \multicolumn{3}{c}{\textbf{Cross Platform}} & \multirow{2}*{\parbox{1.5cm}{\centering\textbf{Computation\\Accelerating}}} & \multirow{2}*{\parbox{1.5cm}{\centering\textbf{Distributed\\Computing}}} & \multirow{2}*{\parbox{1.5cm}{\centering\textbf{Headless\\Simulation}}} \\ \cmidrule{2-7}
        & Disk space & Memory & GPU & Windows & Linux & MacOS & & \\ \midrule[1pt]
        SUMO & 512 MB & 128 MB & - & $\surd$ & $\surd$ & $\surd$ & - & - & $\surd$ \\ \midrule
        Webots & 512 MB & 2 GB  & 512 MB & $\surd$ & $\surd$ & $\surd$ & $\surd$ & - & $\surd$  \\ \midrule
        VDrift & 1 GB & 512 MB & - & $\surd$ & $\surd$ & $\surd$ & - & - & - \\
        CommonRoad & 32 MB & 2 GB & - & $\surd$ & $\surd$ & $\surd$ & - & - & - \\
        highway-env & 1 MB & 128 MB & - & $\surd$ & $\surd$ & $\surd$ & - & - & - \\
        SMARTS & 8 GB & 2 GB & - & $\surd$ & $\surd$ & $\surd$ & - & $\surd$ & - \\
        MetaDrive & 1.5 GB & 3GB & 1GB & $\surd$ & $\surd$ & $\surd$ & - & $\surd$ & $\surd$ \\
        NuPlan & 7 GB & 128 MB & -& $\surd$ & $\surd$ & $\surd$ & - & - & - \\
        InterSim & 512 MB & 256 MB & - & $\surd$ & $\surd$ & $\surd$ & - & - & - \\
        TBSim & 5 GB & 7 GB & - & $\surd$ & $\surd$ & $\surd$ & - & - & - \\
        Waymax & 3 GB & 128 MB & - & $\surd$ & $\surd$ & $\surd$ & $\surd$ & $\surd$ & - \\ \midrule
        CARLA & 20 GB & 8 GB & 8GB & $\surd$ & $\surd$ & - & - & - & $\surd$  \\ \bottomrule[2pt]
    \end{tabular}
\end{table*}

\subsubsection{Improvement suggestions}

The efficiency of the simulators is closely related to the software architecture and code quality. It is crucial to make thoughtful choices of the back-end physics engine and programming language at the design stage. Some physics engines come with rendering library dependencies that are incompatible with headless and computational functionalities. Moreover, choosing a high-level programming language could slow down computational speed. Implementing a Continuous Integration (CI) workflow is imperative during the development process, which automatically checks performance and promptly identifies coding issues, ensuring a smoother and more reliable development pipeline.

\section{Conclusion}
\label{conclusion}

This paper presents a comprehensive overview of the history of simulators for autonomous driving over the past three decades. According to the increment pattern of simulators, the timeline is divided into three distinct phases: the incipient period (1990s-2000s), the dormant period (2000s-2015), and the outbreak period (2015-present). A tendency toward open-source accessibility and comprehensive functionality is revealed in the evolution process.

After investigating the influential commercial and open-source simulators, they are classified into four task-specific classes: traffic flow simulator, sensory data simulator, driving policy simulator, and vehicle dynamics simulator. Additionally, simulators that enclose multiple functionalities are grouped as the fifth type, comprehensive simulators.

Through a systematic exploration of simulators, this paper identifies critical issues with a particular focus on open-source variants. Key concerns include simulation fidelity, efficiency, and data format inconsistency. Fidelity issues extend to the synthesis of sensory data, traffic scenarios, and vehicle dynamics. In generating traffic scenarios, efficiency challenges arise in constructing detailed maps with rich traffic information with minimal human labor, intricately linked to data format inconsistency. Simulation efficiency is a global concern. The potential hazards posed by these issues to the validity of ADS models developed in simulators are justified, serving as a reminder to simulator users. The paper also reviews potential mitigation methods for these impacts.

This paper is presented as a comprehensive guide for selecting an appropriate simulator for ADS research tasks. Simultaneously, it serves as a call to address the pressing limitations present in actively maintained open-source simulators. It is believed that advancements in simulation technology have the potential to accelerate the implementation and deployment of autonomous driving technology.

\bibliographystyle{IEEEtran}
\bibliography{
    ref
}

\begin{IEEEbiography}[{\includegraphics[width=1in,height=1.25in,clip,keepaspectratio]{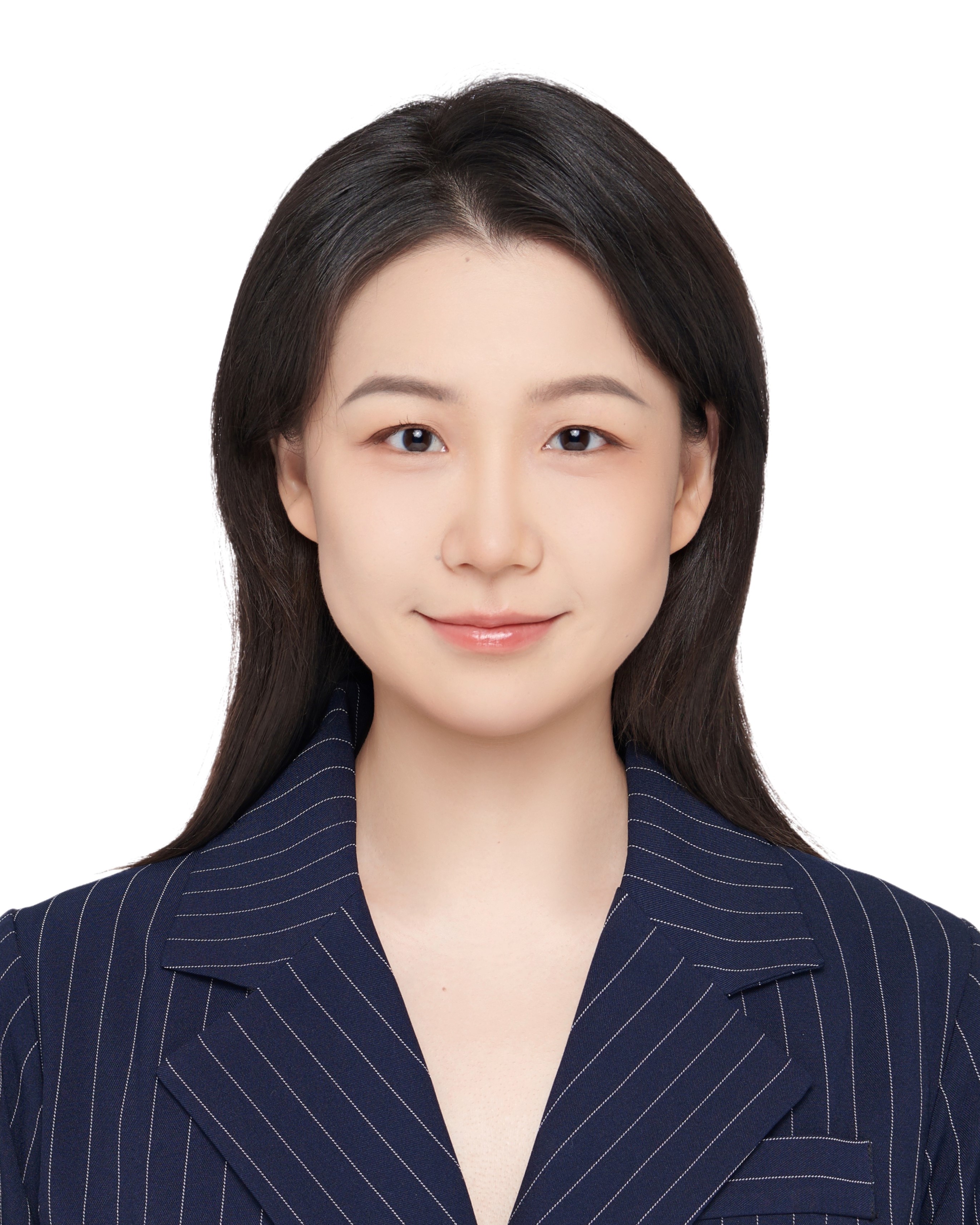}}]{Yueyuan LI} 
    received a Bachelor's degree in Electrical and Computer Engineering from the University of Michigan-Shanghai Jiao Tong University Joint Insitute, Shanghai, China, in 2020. She is working towards a Ph.D. degree in Control Science and Engineering from Shanghai Jiao Tong University.
    
    Her main fields of interest are the security of the autonomous driving system and driving decision-making. Her current research activities include driving decision-making models, driving simulation, and virtual-to-real model transferring.
\end{IEEEbiography}

\begin{IEEEbiography}[{\includegraphics[width=1in,height=1.25in,clip,keepaspectratio]{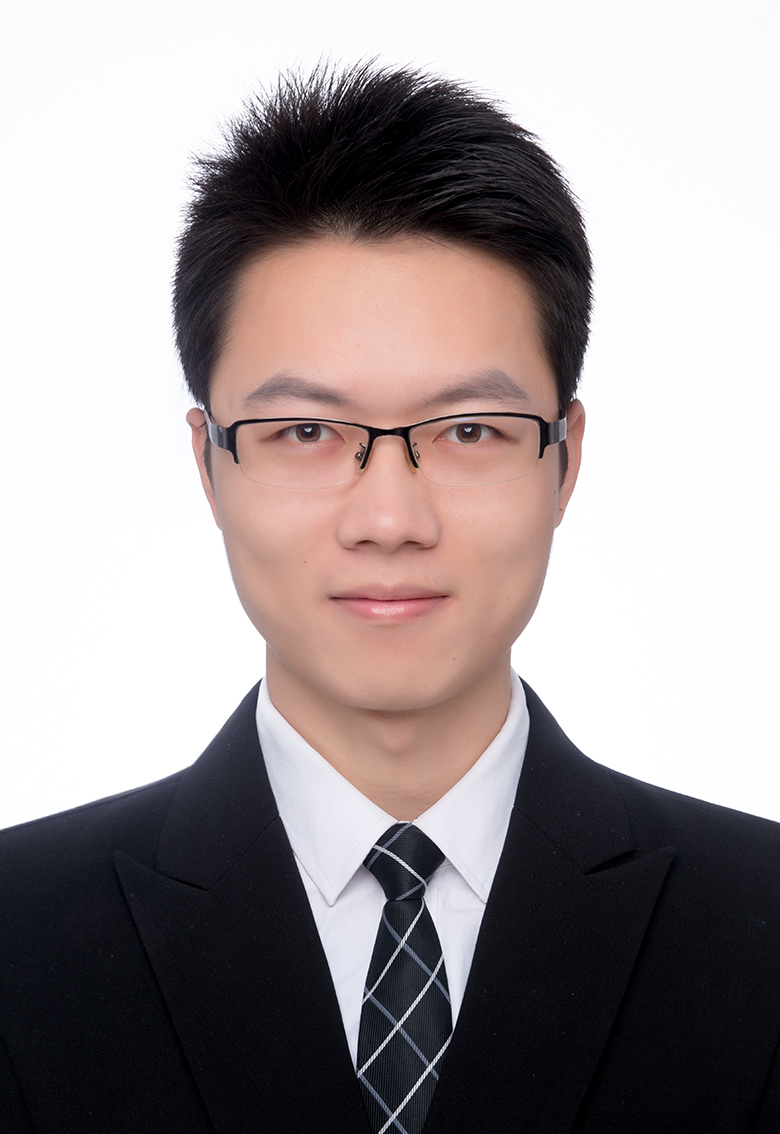}}]{Wei YUAN}    
    received his Master’s and Ph.D. degrees in Automation from Shanghai Jiao Tong University, Shanghai, China, in 2017 and 2021, respectively. Presently, he is a postdoctoral researcher at Shanghai Jiao Tong University.
    
    His main fields of interest are autonomous driving systems, computer vision, deep learning, and vehicle control. His current research activities include end-to-end learning-based vehicle control and decision-making.
\end{IEEEbiography}

\begin{IEEEbiography}[{\includegraphics[width=1in,height=1.25in,clip,keepaspectratio]{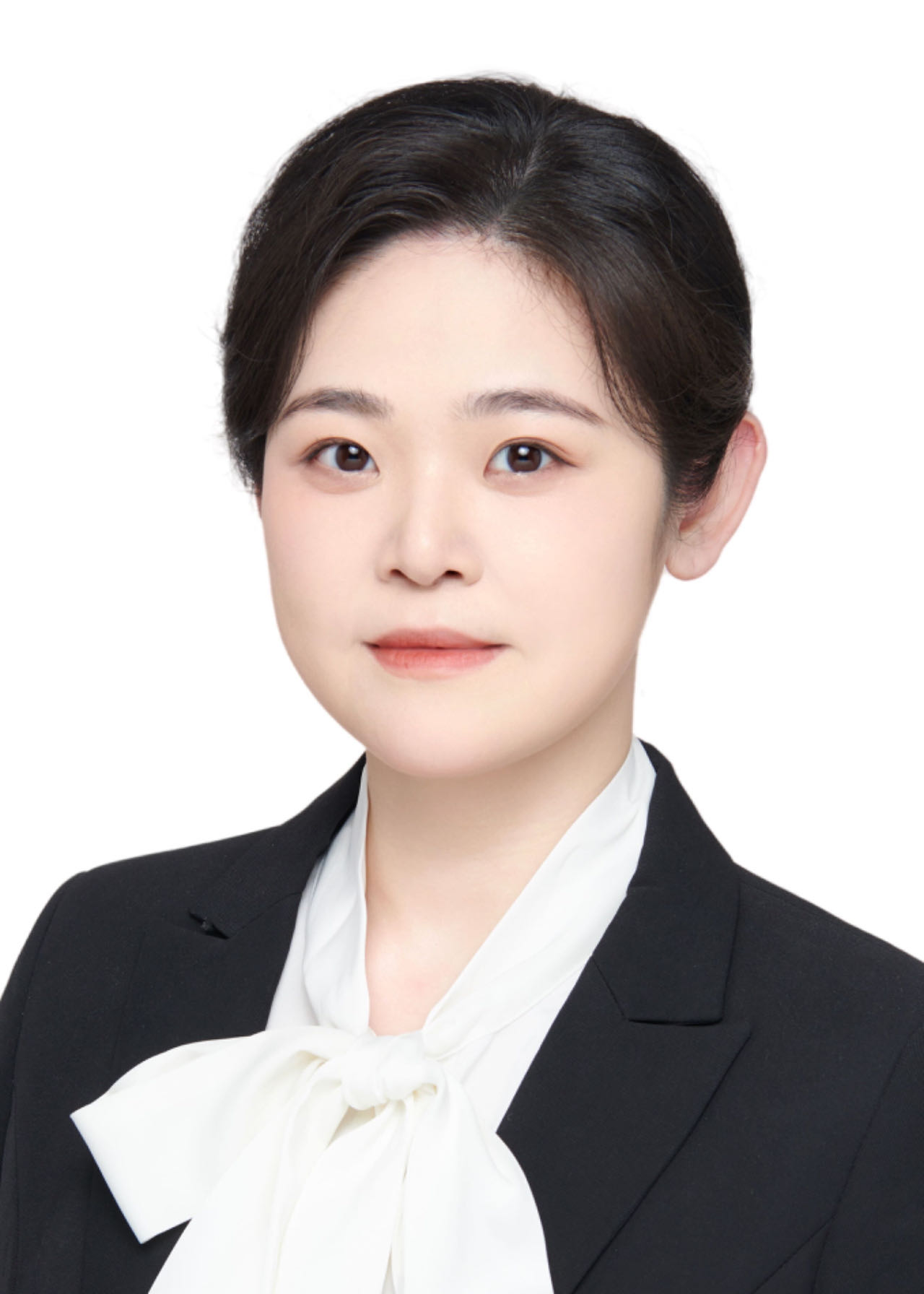}}]{Songan Zhang}
    received B.S. and M.S. degrees in automotive engineering from Tsinghua University in 2013 and 2016, respectively. Then, she went to the University of Michigan, Ann Arbor, and got a Ph.D. in mechanical engineering in 2021. She is currently working at Ford Motor Company in the Robotics Research Team. Her research interests include accelerated evaluation of autonomous vehicles, model-based reinforcement learning, and meta-reinforcement learning for autonomous vehicle decision-making.    
\end{IEEEbiography}

\begin{IEEEbiography}[{\includegraphics[width=1in,height=1.25in,clip,keepaspectratio]{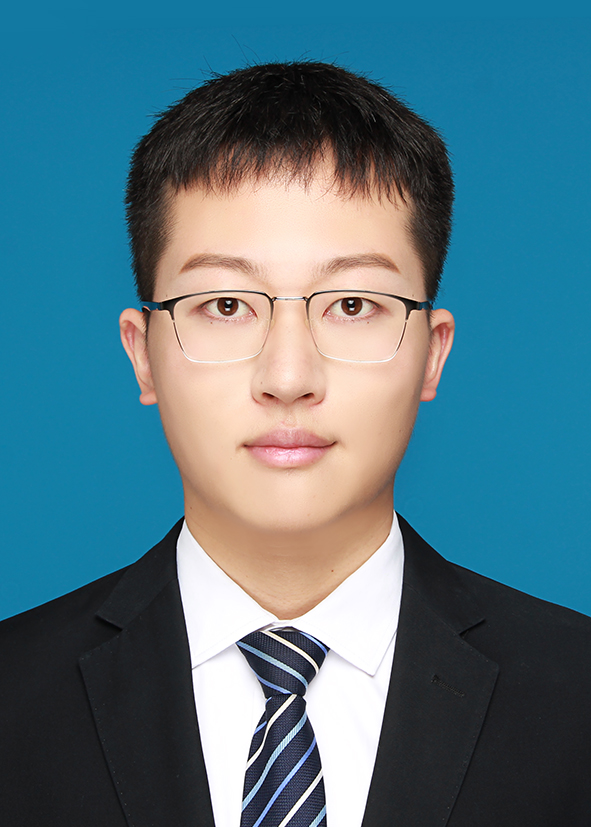}}]{Weihao YAN}
    received a Bachelor's degree in Automation from Shanghai Jiao Tong University, Shanghai, China, in 2020. He is working towards a Ph.D. degree in Control Science and Engineering from Shanghai Jiao Tong University.
    
    His main fields of interest are autonomous driving systems, computer vision, and domain adaptation. His current research activities include virtual-to-real transfer learning, scene segmentation, and foundation models.
 
\end{IEEEbiography}

\begin{IEEEbiography}[{\includegraphics[width=1in,height=1.25in,clip,keepaspectratio]{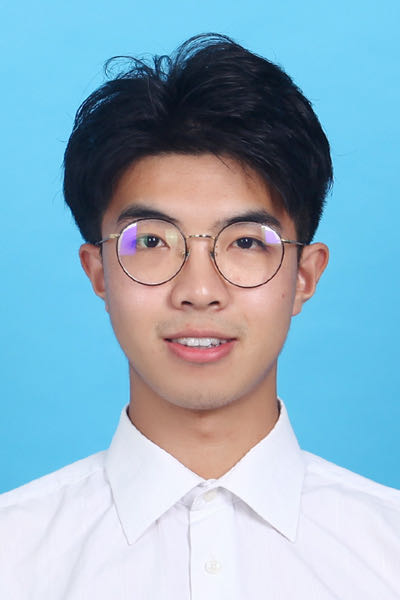}}]{Qiyuan SHEN}
    received a Bachelor's degree in Automation from Northeastern University, Shenyang, China, in 2022. He is currently working towards a Ph.D. degree in Control Science and Engineering at Shanghai Jiao Tong University.
    
    His main fields of interest are the localization of the autonomous driving system and SLAM. His current research activities include multi-modal mapping, cross-modal localization, and geometrical calibration for vision and LiDAR.

\end{IEEEbiography}

\begin{IEEEbiography}[{\includegraphics[width=1in,height=1.25in,clip,keepaspectratio]{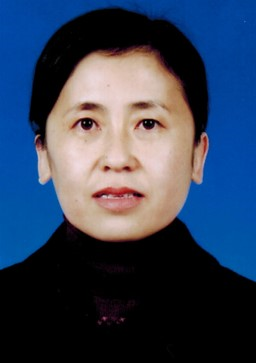}}]{Chunxiang WANG}
    received a Ph.D. degree in Mechanical Engineering from Harbin Institute of Technology, China, in 1999. She is currently an associate professor in the Department of Automation at Shanghai Jiao Tong University, Shanghai, China. 
    
    She has been working in the field of intelligent vehicles for more than ten years and has participated in several related research projects, such as European CyberC3 project, ITER transfer cask project, etc. Her research interests include autonomous driving, assistant driving, and mobile robots. 
\end{IEEEbiography}

\begin{IEEEbiography}[{\includegraphics[width=1in,height=1.25in,clip,keepaspectratio]{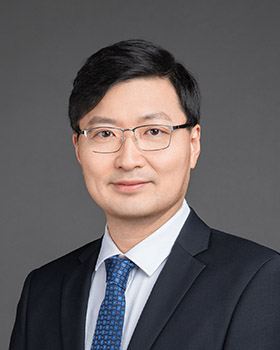}}]{Ming YANG}
    received his Master’s and Ph.D. degrees from Tsinghua University, Beijing, China, in 1999 and 2003, respectively. Presently, he holds the position of Distinguished Professor at Shanghai Jiao Tong University, also serving as the Director of the Innovation Center of Intelligent Connected Vehicles. Dr. Yang has been engaged in the research of intelligent vehicles for more than 25 years.
\end{IEEEbiography}

\vfill

\end{document}